\documentclass[journal]{IEEEtran}

\usepackage{url}  
\usepackage{graphicx}  
\usepackage{amsmath,amssymb}
\usepackage{color}
\usepackage{epstopdf}
\usepackage{array}
\usepackage{multirow}
\usepackage{bbm}
\usepackage{amsfonts}
\usepackage{booktabs}
\usepackage{bm}
\usepackage{float}
\usepackage{subfigure}
\usepackage{cite}
\usepackage{ifpdf}
\usepackage[switch]{lineno}
\usepackage[ruled, lined, longend, linesnumbered]{algorithm2e}

\begin{document}
\title{Fine-Grained Image Captioning \\ with Global-Local Discriminative Objective}

\author{Jie Wu, Tianshui Chen, Hefeng Wu, Zhi Yang, Guangchun Luo, and Liang Lin
\thanks{
This work was supported in part by the National Key Research and Development Program of China under Grant No. 2018YFC0830103, in part by National Natural Science Foundation of China (NSFC) under Grant No. 61876045 and U1811463, in part by National High Level Talents Special Support Plan (Ten Thousand Talents Program), in part by the Natural Science Foundation of Guangdong Province under Grant No. 2017A030312006, and in part by Zhujiang Science and Technology New Star Project of Guangzhou under Grant No. 201906010057. 

Jie Wu, Hefeng Wu and Zhi Yang are with Sun Yat-Sen University, Guangzhou, China.
Tianshui Chen and Liang Lin are with Sun Yat-Sen University and DarkMatter Research.
Guangchun Luo are with University of Electronic Science and Technology of China, Chengdu, China.

Corresponding author is Tianshui Chen (Email: tianshuichen@gmail.com).}
}

\markboth{IEEE Transactions on Multimedia}%
{J. Wu \MakeLowercase{\textit{et al.}}: Fine-Grained Image Captioning with a Global-Local Discriminative Objective}

\maketitle

\begin{abstract}
Significant progress has been made in recent years in image captioning, an active topic in the fields of vision and language. However, existing methods tend to yield overly general captions and consist of some of the most frequent words/phrases, resulting in inaccurate and indistinguishable descriptions (see Figure \ref{fig:motivation}). This is primarily due to (i) the conservative characteristic of traditional training objectives that drives the model to generate correct but hardly discriminative captions for similar images and (ii) the uneven word distribution of the ground-truth captions, which encourages generating highly frequent words/phrases while suppressing the less frequent but more concrete ones. In this work, we propose a novel global-local discriminative objective that is formulated on top of a reference model to facilitate generating fine-grained descriptive captions. Specifically, from a global perspective, we design a novel global discriminative constraint that pulls the generated sentence to better discern the corresponding image from all others in the entire dataset.
From the local perspective, a local discriminative constraint is proposed to increase attention such that it emphasizes the less frequent but more concrete words/phrases, thus facilitating the generation of captions that better describe the visual details of the given images. We evaluate the proposed method on the widely used MS-COCO dataset, where it outperforms the baseline methods by a sizable margin and achieves competitive performance over existing leading approaches. We also conduct self-retrieval experiments to demonstrate the discriminability of the proposed method.
\end{abstract}


\IEEEpeerreviewmaketitle

\section{Introduction}

\IEEEPARstart{I}{mage} captioning, i.e., automatically generating descriptive sentences of images, has received increasing attention in the fields of vision and language in recent years. Compared with other image semantic analysis tasks such as object detection\cite{zhu2019attention,chen2016disc} or fine-grained image recognition\cite{peng2018object,chen2018knowledge}, image captioning provides a deeper and more comprehensive understanding of the images and extends to a wide range of applications, including image retrieval, scene graph generation and video captioning.

 \begin{figure}[t]
\centering
\includegraphics[width=0.98\linewidth]{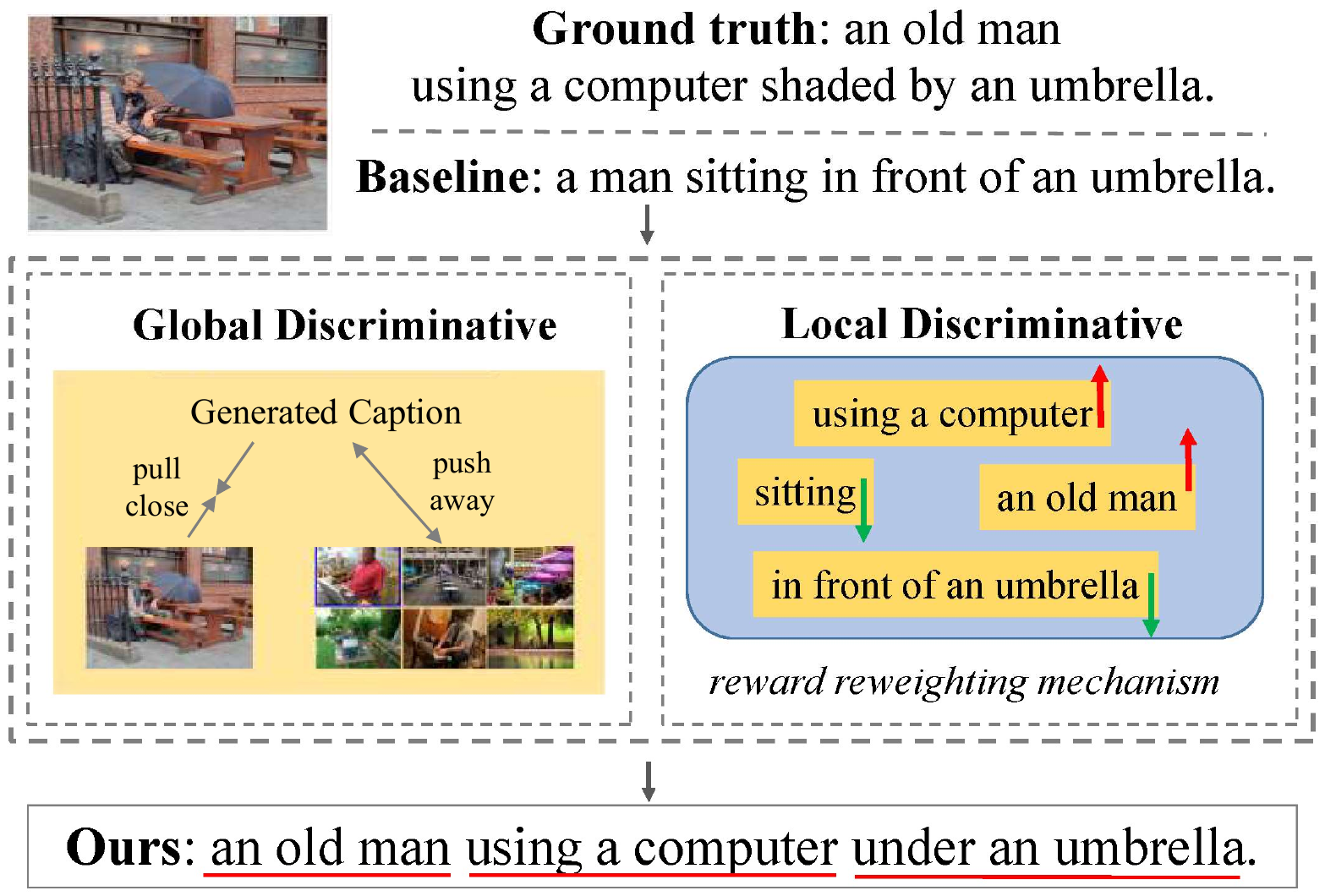}
\caption{An example of how our proposed method aids in generating more fine-grained and accurate descriptions. The proposed method consists of a global discriminative objective and a local discriminative objective. The local discriminative constraint employs a reward reweighting mechanism to increase the rewards of some informative phrases (such as ``using a computer'' and ``an old man'') and decrease the rewards of some inaccurate or universal phrases (such as ``sitting'' and ``in front of an umbrella''). }
\label{fig:motivation}
\end{figure}

With the advancement of deep learning, the existing approaches \cite{rennie2017self,gu2017stack} generally employ a neural-network-based encoder-decoder architecture \cite{vinyals2015show} and resort to a reinforcement learning method \cite{williams1992simple} to optimize this task. Despite acknowledged successes, the captions generated by these leading methods \cite{gu2017stack,anderson2018bottom} are often overly rigid and tend to repeat the words/phrases that frequently appear in the training set. Thus, these captions can hardly describe the corresponding images with the desired accuracy. The reasons are primarily twofold: (1) the conservative characteristic of traditional training objectives (e.g., maximum likelihood estimation (MLE) or consensus-based image description evaluation (CIDEr)), which tends to encourage generating overly universal captions that in turn are hard to be used to discriminate similar images; and (2) the uneven word distribution of ground-truth captions, which encourages generating highly frequent words/phrases while suppressing the less frequent but more concrete ones. For example, given an image shown in Figure \ref{fig:motivation}, existing methods are inclined to generate the high-frequency and common phrase ``a man'' that provides little discriminative or informative cues. Obviously, the less frequent phrase ``an old man'' is a more accurate choice. Another issue is that some images share similar contents, and these methods driven by conservative training objectives tend to pay more attention to such contents, thus resulting in similar or even identical captions for such images. In recent years, some works have generated diverse and discriminative captions with the help of adversarial learning \cite{shetty2017speaking,dai2017towards} or ranking loss \cite{luo2018discriminability}. However, these methods focus more on diversity and may not balance diversity and accuracy well.

In this work, we propose a novel global-local discriminative objective to train the captioning model.
From the global perspective, we first design a global discriminative constraint that encourages the generated caption to better describe the corresponding image rather than other similar images. To this end, we employ the ranking loss that pulls the generated caption to match the corresponding image while pushing the caption away from other similar images. From a local perspective, we develop a local discriminative constraint that pays more attention to finer-grained words/phrases. As suggested in \cite{vedantam2015cider}, the fine-grained words/phrases generally occur less frequently because they merely describe some distinct and detailed contents of some specific images. Thus, we implement this constraint by assigning higher rewards to less frequent words. This can well address the uneven word distribution issue and help to capture more informative visual details of the images. The two constraints are based on the reference model to facilitate generating discriminative captions while simultaneously improving the accuracy. An example of how our proposed method aids in generating finer-grained and more accurate description is presented in Figure \ref{fig:motivation}.
There is a certain correlation between fine-grained and accurate. When our model generates accurate sentences, it actually reflects more fine-grained details at the same time. Similarly, the discriminative captions can also improve the accuracy of the caption. It can be regarded as a process of mutual promotion.

 The main contributions of this work are summarized as follows. First, we propose a new global-local discriminative objective to optimize the image captioning model, which encourages generating fine-grained and discriminative descriptions from both global and local perspectives. This objective can readily be applied to improve existing captioning models. Second, the proposed global discriminative constraint incorporates a reliable term that enables stable and effective training. Third, our local discriminative constraint establishes a novel word-level content-sensitive reward function. Moreover, we conduct extensive experiments on the widely used MS-COCO dataset, which demonstrates that our approach outperforms the baseline methods by a sizable margin and achieves competitive performance over existing leading approaches.
We also perform self-retrieval experiments \cite{dai2017contrastive} and prove that our proposed method exhibits superior discriminability over existing leading and baseline methods. \emph{We have released our codes and trained models at \url{https://github.com/HCPLab-SYSU/Fine-Grained-Image-Captioning}.}

\section{Related Work}
Recently, image captioning \cite{cho2015describing} has been a popular research task in the field of artificial intelligence, which attempts to generate natural language sentences that describe the visual contents. In real life, image captioning has a great impact, for instance, by aiding visually impaired users in understanding the visual surroundings around them. Recent progress in image captioning has also motivated the exploration of its applications for video captioning \cite{yang2018video} and question answering \cite{xue2018better,xue2017unifying}.

Recent advances in image captioning have benefited from the encoder-decoder pipeline that adopted a convolutional neural network (CNN) \cite{he2016deep} to encode a semantic image representation and a recurrent neural network (RNN) \cite{hochreiter1997long} to decode the representation into a descriptive sentence. As a pioneering work, \cite{vinyals2015show} simply used a GoogLeNet \cite{ioffe2015batch} model to extract an image feature, which was then fed into a long short-term memory (LSTM) network to generate the sentence. Additionally, attention mechanisms \cite{bahdanau2014neural} have recently enjoyed widespread use in computer vision tasks \cite{du2018recurrent,wang2017multi,chen2018recurrent,chen2019learning,chen2018fine}.
Some researchers introduced attention mechanisms \cite{xu2015show,lu2017knowing,gu2017stack,tan2019comic} or learn more discriminative image features \cite{dong2018predicting,yan2019stat} and thus substantially improved the captioning performance. For example, \cite{xu2015show} further learned to locate attentional regions that were highly related to the semantic content to help prediction.
\cite{yao2018exploring,yang2019auto} managed to explore visual relationship \cite{chen2019knowledge} between objects in the graph convolution network for generating accurate captions.

MLE, which aims to maximize the likelihood of the ground-truth word at time step $t$ given the ground-truth words of the previous $t-1$ time steps,  has traditionally been applied to optimize the captioning models \cite{xu2015show}. However, these models suffered from an exposure bias issue, resulting in poor captioning performance \cite{ranzato2015sequence}.
To address these issues, recent works \cite{ranzato2015sequence,rennie2017self,gao2019self} introduced the policy-gradient-based reinforcement learning technique \cite{williams1992simple} for sequence-level training, which directly optimized the discrete and non-differentiable evaluation metrics for the tasks. For example, Ranzato et al. \cite{ranzato2015sequence} defined the reward based on a sequence-level metric (e.g., bilingual evaluation understudy (BLEU) \cite{papineni2002bleu} or recall-oriented understudy for gisting evaluation (ROUGE) \cite{lin2004rouge}) that was used as an evaluation metric during the test stage to train the captioning model, thus leading to a notable performance improvement.
Similarly, Zhang et al. \cite{zhang2017actor} designed an actor-critic algorithm that formulated a per-token advantage function and value estimation strategy into the reinforcement-learning-based captioning model to directly optimize non-differentiable quality metrics of interest.  Rennie et al. \cite{rennie2017self} proposed a self-critical sequence training approach that normalized the rewards using the output of its own test-time inference algorithm for steadier training. Chen et al. \cite{chen2017temporal} introduced the temporal-difference (TD) learning method to further account for the correlation between temporally successive actions.

Although these methods have achieved impressive successes over the past several years, they tend to generate overly rigid sentences that are generally composed of the most frequent words/phrases, leading to inaccurate and indistinguishable descriptions.
Zhang et al. \cite{zhang2018high} created a mechanism of fine-grained and semantic-guided visual attention to generate captions of high accuracy, completeness, and diversity. This attention mechanism can accurately correlate relevant visual information with each semantic in the text.
In addition, a series of efforts were devoted to exploring the generation of diverse and discriminative descriptions because diversity and discriminability are also considered to be important properties for the generated captions \cite{dai2017towards,luo2018discriminability}.
Motivated by adversarial learning \cite{NIPS2014_5423}, existing methods \cite{dai2017towards,shetty2017speaking,yang2018video} also adapted this technology to generate humanoid and natural captions.
 For instance, \cite{dai2017towards} developed a conditional generative adversarial network (CGAN) \cite{mirza2014conditional} framework that jointly learned a generator to produce descriptions conditioned on images and an evaluator to assess how good and natural the generated captions were. Shetty et al. \cite{shetty2017speaking} adopted adversarial training to enable the distribution of the generated caption to better match that of humans. Although these methods could generate diverse and human-like sentences, they primarily focused on diversity and naturalness, and they suffered from a performance drop on evaluation metrics such as CIDEr \cite{vedantam2015cider} and BLEU \cite{papineni2002bleu}.

\begin{figure*}[t]
\centering
\includegraphics[width=0.95\linewidth]{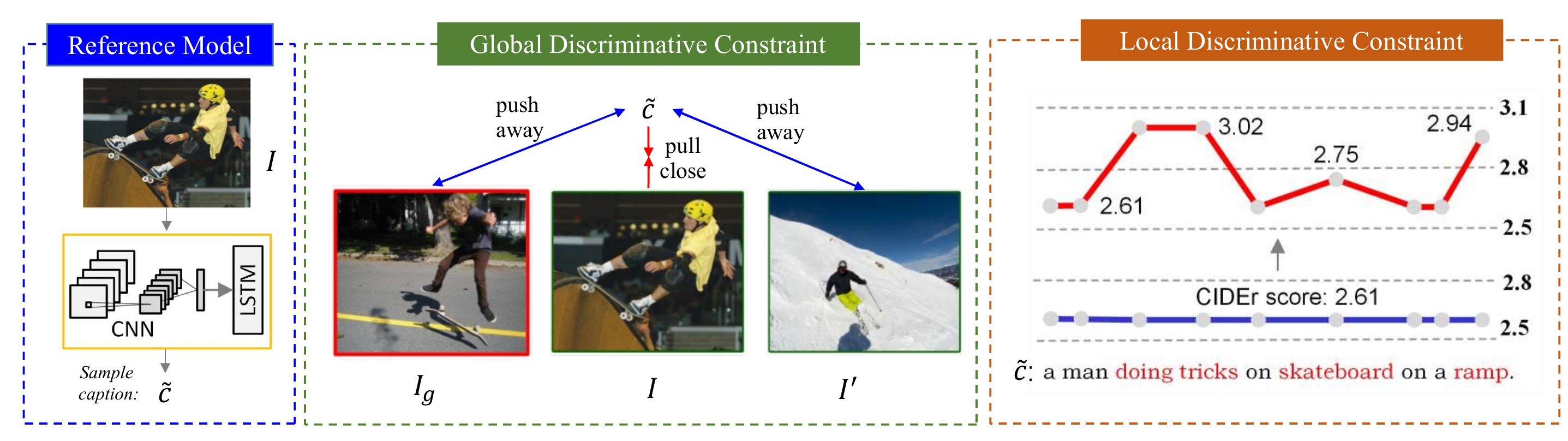}
\caption{An illustration of the proposed objective. It consists of a global discriminative constraint and a local discriminative constraint that is formulated on top of a reference model to encourage generating more accurate and fine-grained descriptions.}
\label{fig:model}
\end{figure*}

 Ranking algorithms \cite{faghri2017vse++,chen2016deep} were designed to pull the matched instances into close proximity with each other and to push the unmatched instances to increase their distance from each other \cite{chen2016deep}.
A series of works applied these algorithms to facilitate the diversity of various generation tasks, including visual question answering \cite{goyal2017making}, image generation \cite{saquil2018ranking,diba2017object}, and video forecasts \cite{xiong2018learning,yang2018text2video}. For instance, Xiong et al. \cite{xiong2018learning} proposed a multistage dynamic generative adversarial network and designed an adversarial ranking loss to optimize this network to encourage the generated video to be closer to the real one while being farther away from the input video. To enable generating diverse and discriminative sentences, recent works \cite{luo2018discriminability,dai2017contrastive} also formulated the ranking loss as an additional constraint on top of current captioning models.
\cite{luo2018discriminability} introduced hard negative triplet loss \cite{schroff2015facenet} as an extra constraint to train the captioning model such that , being able to generate more discriminative captions.
However, this kind of method may lead to unstable training and model degradation. The reason is that it just measures discriminability among samples of a minibatch during training, but the reward built on the minibatch can be invalid in the cases in which the images of the minibatch are not similar to some extent. More seriously, the captioning model will be misled and presume that the generated caption is discriminative. Such cases occur more often when the size of the minibatch becomes smaller. Consequently, it required training with a relatively large minibatch size to ensure discriminability during training.

In contrast to the aforementioned works, our method is able to overcome the above problem by improving the discriminability from a totally global perspective. The proposed global discriminative constraint incorporates a term that uses the most similar image in the whole dataset to provide significant distinctive rewards. This term serves as a basic and reliable reward to enable stable and effective training.
Furthermore, our method further introduces a local discriminative constraint, which pays more attention to the less frequent words and encourages describing more detailed and fine-grained content of the input image. In this way, our method can generate discriminative captions and simultaneously enhance the overall performance on evaluation metrics.

\section{Methodology}
\subsection{Overview}
Currently, advanced and typical image captioning methods adopt the encoder-decoder framework and generally resort to the reinforcement learning (RL) method for optimization.
In this work, we also utilize this encoder-decoder pipeline \cite{xu2015show} as our reference model. Specifically, the pipeline involves a CNN \cite{he2016deep} to encode the input image $I$ into a semantic feature representation and an LSTM network \cite{hochreiter1997long} to decode this representation into the target descriptive sentence $c$. This process can be formulated as
\begin{align}
v = \phi(I); \quad  c=\psi(v),
\end{align}
where $\phi$ denotes the CNN encoder and $\psi$ represents the LSTM decoder.
During training, the sequential word generation process is formulated as a sequential decision-making problem, and the RL method is introduced to learn a policy network for decision making.
Let $\theta$ denote the parameters of the captioning model and $p_\theta=p(c|I;\theta)$ be the conditional distribution over captions.
Then, RL commonly aims to minimize the negative expected reward, formulated as
\begin{align}
 \mathcal{L}(\theta)=-\mathbb{E}_{\tilde{c} \sim p_\theta}[R_{C}(\tilde{c})],
\end{align}
where $\tilde{c}=\{w^s_1, \dots, w^s_t, \dots, w^s_T\}$ is a caption sampled from the conditional distribution $p_\theta$ (i.e., $\tilde{c} \sim p_\theta$), and $w^s_t$ is the word at time step $t$ in $\tilde{c}$.
 $R_{C}(\tilde{c})$ is a reward defined for the caption $\tilde{c}$.

Training with the reinforcement learning algorithm requires designing an appropriate reward function.
Currently, the reward $R_{C}(\tilde{c})$ is often defined based on the CIDEr score \cite{vedantam2015cider} because it can well measure the quality of the generated captions. However, this reward can hardly enable generating discriminative captions for similar images. Moreover, most existing works use $R_{C}(\tilde{c})$ to provide the same caption-level reward for each word, i.e.,
\begin{align}
 R(w^s_t)=R_{C}(\tilde{c}), \quad t=1,...,T
\end{align}
which is contrary to the appropriate credit assignment. Hence, this setting is susceptible to the uneven word distribution that encourages generating highly frequent words/phrases while suppressing the less frequent but more fine-grained ones.

To solve these issues, we design a global-local discriminative objective for the reward, which is formulated as two constraints based on the above-described reference model, as shown in Figure \ref{fig:model}.
Concretely, to encourage the generated captions to describe the corresponding images well, we develop a global discriminative constraint that pulls the generated caption to match the corresponding image while pushing the caption away from other similar images via a ranking loss.
Furthermore,the local discriminative constraint provides higher rewards for the less frequent but fine-grained words via a word-level reward reweighting mechanism instead of treating all words equally. In this way, the model will pay more attention to these words and thereby alleviate the strong bias of the generated words/phrases.
Therefore, the reward $R(w^s_t)$ can be defined as
\begin{equation}
R(w^s_t) = R_{\mathrm{GD}}(I, \tilde{c}) +  R_{\mathrm{LD}}(w^s_t),
\end{equation}
where $R_{\mathrm{GD}}$ and $R_{\mathrm{LD}}$ are the two rewards defined according to the global-local discriminative constraints, respectively. We introduce these two rewards in detail in the next subsections. Consequently, we aim to minimize the following objective:
\begin{align}\label{eq:sumloss}
 \mathcal{L}(\theta)=-\mathbb{E}_{\tilde{c} \sim p_\theta}[\sum_{t=1}^{T}R(w^s_t)].
\end{align}

\subsection{Global Discriminative Constraint}
Discriminability is essential for fine-grained image captioning. Some works \cite{vedantam2017context, mao2016generation} focus on designing different loss functions to generate discriminative sentences. In this paper, we design a global discriminative constraint that resorts to the visual-semantic embedding module \cite{faghri2017vse++} to act as an evaluator to measure the uniqueness of captions. This constraint is designed to pull the generated captions to better match the corresponding image rather than the others.
To this end, we first introduce a function $s(I,c)$ that measures the similarity of an image $I$ and sentence $c$. The detail of this score function will be described in Section \ref{sec:setting}. Then, given an input image $I$ and its sampled caption $\tilde{c}$, it is expected that the score $s(I,\tilde{c})$ is higher than score $s(I_a,\tilde{c})$ for any image $I_a$ taken from the training image set $\mathcal{I}$. Here, because it is impractical to compute $s(I_a,\tilde{c})$ for all images during training, we approximate this target by enabling $s(I,\tilde{c})$ to be higher than $s(I_g,\tilde{c})$, in which $I_g$ is the image most similar to $I$, formulated as
\begin{equation}
R_{\mathrm{H}}(I,\tilde{c}) = -[\epsilon +s(I_g,\tilde{c})-s(I,\tilde{c})]_+,
\label{eq:rh}
\end{equation}
where $[x]_+$ is a ramp function defined by $\mathrm{max}(0,x)$. To obtain the most similar image $I_g$ for each image $I$, we extract the image feature using ResNet-101 \cite{he2016deep} pretrained on the ImageNet dataset \cite{russakovsky2015imagenet} and compute the Euclidean distance between features of $I$ and all other images $I_a$. The image with the smallest distance in the entire training set is selected as $I_g$. We can retrieve the most similar image for each image before training, so this process hardly incurs any additional training cost.

\begin{figure*}[t]
\centering
\includegraphics[width=0.95\linewidth]{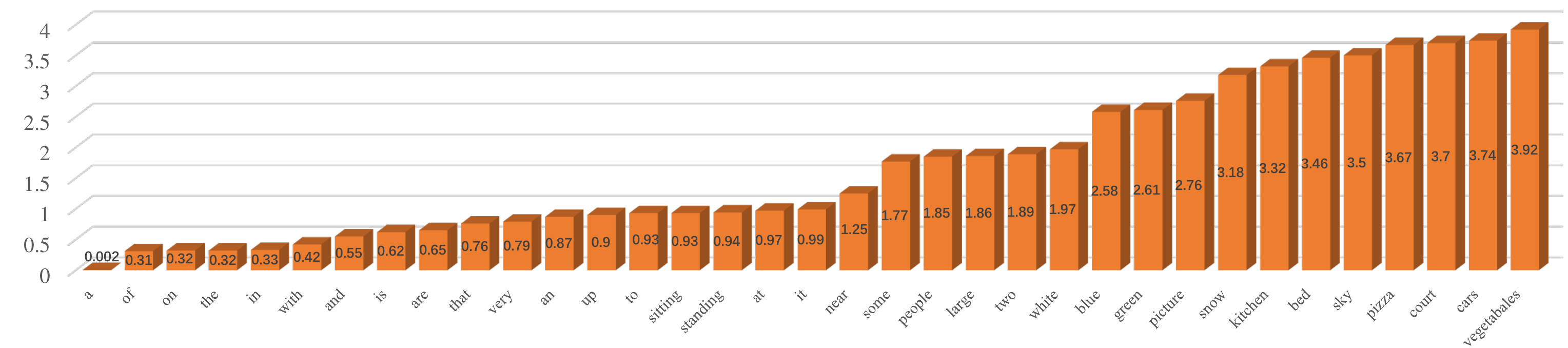}
\caption{The TF-IDF weights for some words (1-gram) in the MS-COCO dataset. Words are sorted by TF-IDF weights.}
\label{fig:1gram}
\end{figure*}

 This reward setting contributes to improving the discriminability from a global perspective, but it merely considers one reference image (i.e., the most similar image). In fact, some other images also share very similar content with the given image. Taking these images into the reward definition manages to further improve the discriminability. Inspired by \cite{luo2018discriminability, liu2018show}, we introduce another ranking target defined on the minibatch during training:
\begin{equation}
\begin{split}
R_{\mathrm{B}}(I,\tilde{c})= -[\epsilon +s(I,c')-s(I,\tilde{c})]_+ \\ -[\epsilon +s(I',\tilde{c})-s(I,\tilde{c})]_+ ,
\end{split}
\label{eq:rb}
\end{equation}
where $I'= \arg\max_{I'\ne I}s(I',\tilde{c})$ is the hardest negative image from the current minibatch and $c'= \arg\max_{c'\ne c}s(I,c)$ is the hardest negative caption in the minibatch.
Thus, $(I, c')$ and $(I', \tilde{c})$ are the hard negative pair defined on the minibatch.
The training data are completely shuffled to select the batch samples in each iteration. Applying sufficient iterations will approximate using the entire training set, so the discriminative loss in minibatches can be considered as part of the global discriminative constraint.
Furthermore, a random minibatch can introduce a high degree of data diversity to facilitate effective training.

Finally, we sum the two terms to obtain the global discriminative reward:
\begin{equation}
R_{\mathrm{GD}}(I,\tilde{c}) = R_{\mathrm{H}}(I,\tilde{c}) + R_{\mathrm{B}}(I,\tilde{c}) .
\label{eq:gd}
\end{equation}


\subsection{Local Discriminative Constraint}
The local discriminative constraint is content-sensitive and expected to assign higher rewards to the words/phrases that concretely describe the visual contents of given images.
Thus, we adopt the reward reweighting mechanism to provide higher rewards to these words/phrases.
In general, some phrases that describe the distinct and detailed contents of some specific images such as ``doing tricks on the ramp'' occur less frequently in the dataset.
To this end, we resort to the term frequency-inverse document frequency (TF-IDF) \cite{robertson2004understanding} weights to compute the frequency of each n-gram (n = 1, 2, 3, 4) phrase in the dataset. Then we adopt a two-stage mechanism to select and reweight the less frequent but informative words.
The two-stage mechanism is designed based on the following assumptions: 1) The fine-grained and detailed phrases are selected according to the computed TF-IDF weights, and these weights will be assigned to each corresponding word and increase their rewards.
2) Some frequently occurring common words, such as ``a'', ``on'' and so forth, should be determined such that their original weights are retained since they are the basic building blocks of almost all sentences.
Below, we describe this mechanism in detail.

In the first stage, we follow \cite{vedantam2015cider} to compute a TF-IDF weight for each n-gram phrase $w_k$ in the sampled sentence $\tilde{c}$:
\begin{equation}
 g_{\omega_k}(\tilde{c})= \frac{n_{\omega_k}(\tilde{c})}{\sum_{\omega \in \Omega}n_{\omega}(\tilde{c})}\mathrm{log}(\frac{|\mathcal{I}|}{\sum _{I_p \in \mathcal{I}} \mathrm{min}(1, \sum_{q}n_{\omega_k}(s_{pq}))}),
\end{equation}
where $\Omega$ is the vocabulary of all n-grams and $\mathcal{I}$ is the set of all images in the dataset. $n_{\omega}(\tilde{c})$ denotes the number of times the n-gram $\omega$ occurs in the sentence $\tilde{c}$. $s_{pq}$ is the $q$-th ground-truth sentence for image $I_p$. The TF-IDF weight $g_{\omega_k}(\tilde{c})$ reflects the saliency of the n-gram $\omega_k$ in the dataset, and a higher weight indicates that this n-gram occurs less frequently across all images in the dataset.
\begin{figure}[!t]
\centering
\includegraphics[width=0.95\linewidth]{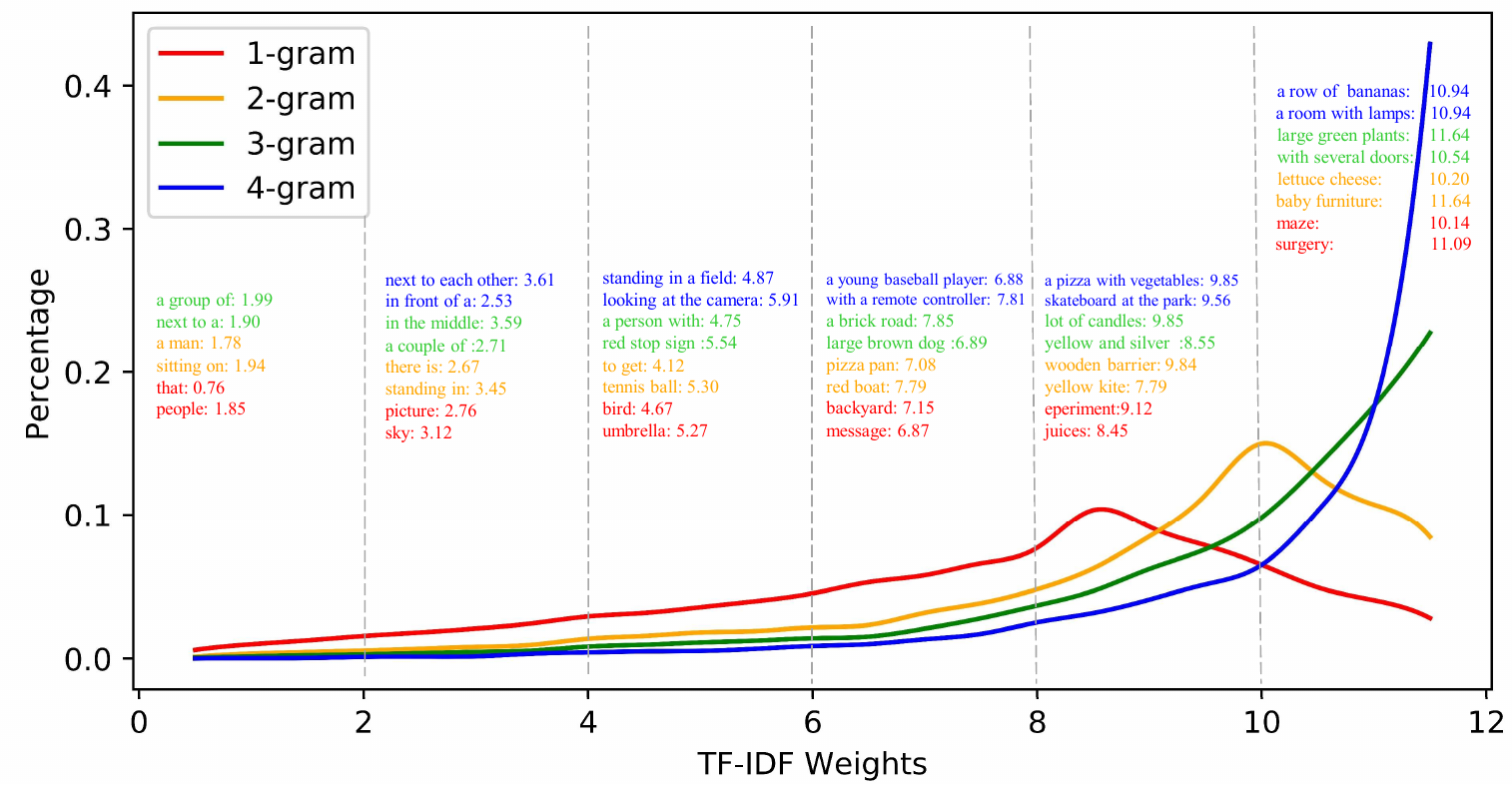}
\caption{The probability distribution and some examples of TF-IDF weights for n-grams (n = 1, 2, 3, 4) in the MS-COCO dataset.}
\label{fig:ngram}
\end{figure}
We statistically show the probability distribution of TF-IDF weights for n-grams in the MS-COCO dataset along with some examples of $n$-grams. The results are summarized in Figure \ref{fig:ngram}.
From Figure \ref{fig:ngram}, we can see that the distribution will concentrate on larger TF-IDF weights with the increase of $n$.
To determine which n-grams should be assigned a higher reward, we introduce a threshold $\lambda$ to select the fine-grained n-grams with TF-IDF weights higher than $\lambda$. This threshold should maintain the informative n-grams, such as ``red stop sign'' and ``a brick road'', while filtering out some frequently occurring n-grams, such as ``a person with'' and ``next to each other''.
Based on the above analysis and the observations in Figure \ref{fig:ngram}, we fix $\lambda$ to 5 in this paper.

Note that not all words in the selected n-gram phrases are informative, particularly some articles and conjunctions. For example, ``in the grass'' is a less frequent $n$-gram, but the article ``the'' and the preposition ``in'' occur frequently in the dataset and are generally less relevant to the image content.

Thus, in the second stage, we utilize the TF-IDF weights to select these words using another threshold $\eta$. These words are sentence-structured words, which should maintain their original rewards. We summarize the TF-IDF weights of some words in Figure \ref{fig:1gram}. As depicted in Figure \ref{fig:1gram}, some less-informative words (such as ``that'' and ``it'') have small TF-IDF weights while some fine-grained words (such as ``blue'' and ``vegetables'') have larger weights. In our work, $\eta$ is set to 1 according to the observations in Figure \ref{fig:1gram}.
More qualitative and quantitative analysis of these two thresholds is summarized in the experimental results.

Finally, we also utilize the computed TF-IDF weights to determine the increase in the reward of each informative word.
The reward definition is inspired by the CIDEr metric \cite{vedantam2015cider}. Concretely, the reward for the local discriminative constraint can be defined as:
\begin{equation}
\begin{split}
 & R_{\mathrm{LD}}(w^s_t)= \\
 & \sum _{w^s_t \in \omega_k} \sum _{j} \frac {\mathrm{min}(g_{\omega_k}(\tilde{c}), g_{\omega_k}(s_{j})) \cdot g_{\omega_k}(s_{j})} {\Vert g_{\omega_k}(\tilde{c}) \Vert \Vert g_{\omega_k}(s_{j})\Vert}  + R_{\mathrm{C}}(\tilde{c}) \\   & \text{if~~} g_{\omega_k}(\tilde{c}) > \lambda \text{~~and~~} g_{w^s_t}(\tilde{c}) > \eta,
\end{split}
\label{eq:LD}
\end{equation}
where $w^s_t$ is the word at time step $t$ in the sampled sentence $\tilde{c}$. $g_{w^s_t}(\tilde{c})$ denotes the 1-gram TF-IDF weight for word $w^s_t$ and $g_{\omega_k}(s_{j})$ denotes the TF-IDF weight for n-grams $\omega_k$ in the ground-truth sentence $s_{j}$.  $s_j$ denotes the $j$-th ground-truth sentence for the input image.
$R_{\mathrm{C}}(\tilde{c})$ is the original caption-level reward \cite{rennie2017self} defined based on the CIDEr score.
The factor $\mathrm{min}(g_{\omega_k}(\tilde{c}), g_{\omega_k}(s_{j}))$ penalizes the condition where specific n-grams are repeated continually until the desired sentence length is achieved.
Equation (\ref{eq:LD}) illustrates the word selection and reward reweighing procedures simultaneously.
$g_{\omega_k}(\tilde{c}) > \lambda$ restrains some less-informative phrases while $g_{w^s_t}(\tilde{c}) > \eta$ removes some common words. If the word does not satisfy these two conditions, the sample word $w^s_t$ will obtain its original reward $R_{\mathrm{C}}(\tilde{c})$.

\begin{figure}[!t]
\centering
\includegraphics[width=0.95\linewidth]{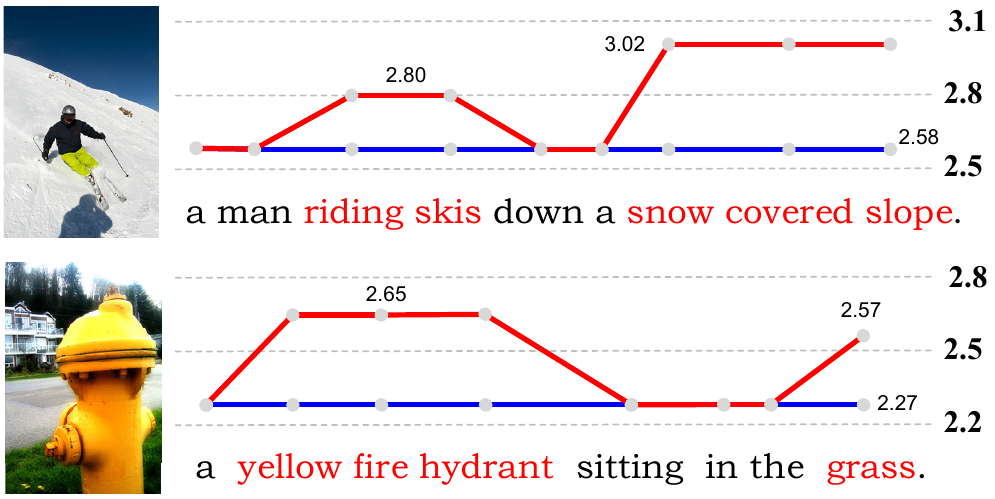}
\caption{The local discriminative reward (the red line) and the traditional CIDEr reward (the blue line).}
\label{fig:cs-reward}
\end{figure}

Figure \ref{fig:cs-reward} illustrates the local discriminative reward for each word in two examples. We find that it is in good agreement with the above-described assumption, in which the article ``a'' and the commonly used word ``man'' maintain the original reward, while the detailed words such as ``slope'' and ``hydrant'' that describe most distinguishable content enjoy higher rewards.
It accurately reflects the relationship between image contents and their rewards.
In this way, the local discriminative constraint is beneficial to provide an appropriate credit assignment for each word.

\subsection{Optimization}
At the training stage, we minimize the objective (\ref{eq:sumloss}) to obtain the caption model. In practice, we utilize a sampling mechanism to approximate the expectation and introduce the algorithm known as REINFORCE \cite{williams1992simple}, \cite{mnih2014recurrent} to compute the gradients, formulated as:
\begin{equation}
\begin{split}
\nabla \mathcal{L}(\theta)&=-\mathbb{E}_{w^s \sim p_\theta}[\sum_{t=1}^{T}R(w^s_t)\nabla\mathrm{log}(p_\theta(w^s_t))] \\                      &=-\frac{1}{M}\sum_{m=1}^{M}\sum_{t=1}^{T}R(w^s_{mt})\nabla\mathrm{log}(p_\theta(w^s_{mt})),
\end{split}
\label{eq:L}
\end{equation}
where $M$ is the number of sampled sentences. The gradient estimated by the above approximation is of high variance, which is not conducive to the convergence of the model. To solve this problem, we follow \cite{chen2017temporal} to introduce a baseline sentence to obtain an unbiased low-variance gradient estimation. Specifically, for each sentence $c_m^s$, we adopt the current model to generate a baseline sentence $c_m^b=\{w^b_{m1}, w^b_{m2}, \dots, w^b_{mT}\}$. We compute the reward of this baseline sentence and normalize the gradient in Equation (\ref{eq:L}) by

\begin{align}
\nabla \mathcal{L}(\theta)&=-\frac{1}{M}\sum_{m=1}^{M}\sum_{t=1}^{T}[R(w^s_{mt})-R(w^b_{mt})]\nabla\mathrm{log}(p_\theta(w^s_{mt})).
\label{eq:gradient}
\end{align}
The difference between $R(w^s_{mt})$ and $R(w^b_{mt})$ is small since the two sentences are both sampled from the same distribution. Thus, the gradient variance is low, leading to more stable updating of parameters in the training process \cite{chen2017temporal}.
Moreover, as shown in Equation (\ref{eq:gradient}), samples with higher rewards will be given larger probability, and the inferior samples will be suppressed.

\begin{table*}[t]
\centering
  \vspace{6pt}
 \caption{Performance (\%) of our proposed and existing state-of-the-art methods on the MS-COCO dataset using the Karpathy test split. We report our results that use the more advanced TDA baseline. ``-'' indicates that the corresponding result is not available. The best and second best results are highlighted in \textcolor{red} {\textbf{red}} and \textcolor{blue} {\underline{blue}} fonts (Best viewed in color).}
\begin{tabular}{c|c|c|c|c|c|c|c|c}
\toprule
\centering Method & BLEU4& BLEU3& BLEU2& BLEU1 & ROUGEL & METEOR &SPICE &CIDEr\\
\hline
GoogleNIC\cite{vinyals2015show} (CVPR2015) &24.6&32.9&46.1&66.6&-&-&-&- \\
MRNN\cite{mao2014deep}  (ICLR2015)&27.0&36.4&50.0&68.0&50.0&23.1&-&86.4 \\
SoftAtt\cite{xu2015show} (ICML2015)&24.3&34.4&49.2&70.7&-&23.9&-&- \\
HardAtt\cite{xu2015show} (ICML2015)&25.0&35.7&50.4&71.8&-&23.0&-&- \\
SemATT\cite{you2016image} (CVPR2016)&30.4&40.2&53.7&70.9&-&24.3&-&- \\
SCACNN\cite{chen2016sca} (CVPR2017)&30.2&40.4&54.2&71.2&52.4&24.4&-&91.2 \\
AdaAtt\cite{lu2017knowing} (CVPR2017)&33.2&44.5&59.1&74.2&-&26.6&-&108.5 \\
Rennie\cite{rennie2017self} (CVPR2017)&34.2&-&-&-&55.7&26.7&-&114.0 \\
MSM\cite{yao2017boosting} (ICCV2017) &32.5&42.9&56.5&73.0&53.8&25.1&-&98.6 \\
ALT-ALTM\cite{ye2018attentive} (TIP2018) &35.5&45.7&59.0&75.1&55.9&27.4&20.3&110.7 \\
TD-ATT\cite{chen2017temporal} (AAAI2018) &34.0&45.6&60.3&76.5&55.5&26.3&-&111.6 \\
Stack-cap \cite{gu2017stack} (AAAI2018) &36.1&47.9&62.5&78.6&\textcolor{blue} {\underline{56.9}}&27.4&20.9&\textcolor{blue} {\underline{120.4}} \\
Up-down\cite{anderson2018bottom} (CVPR2018) &\textcolor{red} {\textbf{36.3}}&-&-&\textcolor{red} {\textbf{79.8}}&\textcolor{blue} {\underline{56.9}}&\textcolor{blue} {\underline{27.7}}&\textcolor{blue} {\underline{21.4}}&120.1 \\
OPR-MCM\cite{zhang2019more} (TIP2019) &35.6&46.0&59.6&75.8&56.0&27.3&-&110.5 \\
N-step SCST\cite{gao2019self} (CVPR2019) &35.0 &46.8 &61.5 &77.9 &56.3 &26.9 & 20.4 &115.2 \\
KMSL\cite{li2019know} (TMM2019) &\textcolor{red} {\textbf{36.3}}&\textcolor{red} {\textbf{48.3}}&\textcolor{red} {\textbf{63.2}}&\textcolor{blue} {\underline{79.2}}&56.8&27.6&\textcolor{blue} {\underline{21.4}}&120.2 \\
TDA+GLD (Ours)  &\textcolor{blue} {\underline{36.1}}&\textcolor{blue} {\underline{48.0}}&\textcolor{blue} {\underline{62.6}}&78.8&\textcolor{red} {\textbf{57.1}}&\textcolor{red} {\textbf{27.8}}&\textcolor{red} {\textbf{21.6}}&\textcolor{red} {\textbf{121.1}}\\
\bottomrule
 \end{tabular}
\label{table:result}
 \end{table*}

\section{Experiments}
In this section, we conduct extensive experiments to compare our method with existing state-of-the-art approaches both quantitatively and qualitatively. We also perform ablation studies to discuss and analyze the contribution of each component of the proposed method.

\subsection{Experimental Settings}\label{sec:setting}
\label{sec:Experiment Settings}
\subsubsection{Datasets} MS-COCO  \cite{chen2015microsoft} is a widely used benchmark for the image captioning task. This dataset contains 123,287 images, with each image annotating five sentences.  Each image in this dataset is equipped with five reference sentences, which describe the images using Amazon Mechanical Turk \cite{rashtchian2010collecting} provided by human annotators. In this work, we follow the Karpathy split \cite{karpathy2015deep} that divides the dataset into a training set of 113,287 images, a validation set of 5,000 images, and a test set of 5,000 images for evaluation. We also submit our results to the online MS-COCO test server (\url{https://www.codalab.org/competitions/3221\#results}) for public comparison with the published methods.

\subsubsection{Evaluation Metrics}
We evaluate our method, the baseline methods and other competitors on widely used metrics, including BLEU \cite{papineni2002bleu}, ROUGEL \cite{lin2004rouge}, METEOR \cite{banerjee2005meteor}, semantic propositional image caption evaluation (SPICE) \cite{anderson2016spice} and CIDEr \cite{vedantam2015cider}. We introduce these metrics in detail as follows.

\noindent\textbf{BLEU }\cite{papineni2002bleu} is defined as the geometric mean of the logarithmic n-gram precision scores. The output is further multiplied by the brevity penalty factor BP to penalize short captions. It is effective for measuring the fraction of n-grams (up to 4-gram) that are in common between a hypothesis and a reference.

\noindent\textbf{ROUGEL }\cite{lin2004rouge} evaluates captions based on the co-occurrence information of n-grams in sentences.  ROUGEL uses the longest common subsequence (LCS)-based F1 score to estimate the LCS of tokens between a hypothesis and a reference.

\noindent\textbf{METEOR }\cite{banerjee2005meteor} is designed to explicitly address the weaknesses in BLEU. It evaluates a translation by computing the harmonic mean of unigram precision and recall based on an explicit
word-to-word matching.

\noindent\textbf{SPICE }\cite{anderson2016spice} takes semantic propositional content into account to assess the quality of image captions. Reference and candidate captions are mapped through dependency parse trees to the semantic scene graphs. Caption quality is determined using an F-score calculated over tuples in the candidate and reference scene graphs.

\noindent\textbf{CIDEr }\cite{vedantam2015cider} is an evaluation metric developed specifically for the task of image captioning.  CIDEr measures the similarity of a generated sentence against a set of ground-truth sentences, capturing the notions of grammaticality, saliency, importance, and accuracy inherently by sentence similarity. CIDEr shows a high agreement with consensus as assessed by humans, and it is widely regarded as the most authoritative metric in this task.

 \begin{table*}[t]
\centering
 \vspace{6pt}
\caption{Performance (\%) of our proposed and existing state-of-the-art methods on the online MS-COCO test server. We report our results using the more advanced TDA baseline. $\dag$ indicates the results of ensemble models (bottom part of the table). The best and second best results are highlighted in \textcolor{red} {\textbf{red}} and \textcolor{blue} {\underline{blue}} fonts (Best viewed in color).}
 \begin{tabular}{c|c|c|c|c|c|c|c|c|c|c|c|c|c|c}
\toprule
\ \multirow{2}*{Method} & \multicolumn{2}{c}{BLEU1} & \multicolumn{2}{c}{BLEU2}& \multicolumn{2}{c}{BLEU3}
 & \multicolumn{2}{c}{BLEU4}& \multicolumn{2}{c}{METEOR} & \multicolumn{2}{c}
 {ROUGEL} & \multicolumn{2}{c}{CIDEr} \\ \cmidrule(lr){2-3} \cmidrule(lr){4-5}
 \cmidrule(lr){6-7}\cmidrule(lr){8-9} \cmidrule(lr){10-11} \cmidrule(lr){12-13} \cmidrule(lr){14-15}
 \ & c5 & c40 & c5 & c40 & c5 & c40 & c5 & c40 & c5 & c40 & c5 & c40 & c5 & c40 \\
 \midrule
SCACNN\cite{chen2016sca} (CVPR2017) &72.5&89.2&55.6&80.3&41.4&69.4&30.6&58.2&24.6&32.9&52.8	&67.2&91.1&92.4\\
MSM\cite{yao2017boosting} (ICCV2017) &75.1 &92.6 &58.8 &85.1 &44.9 &75.1 &34.3 &64.6 &26.6 &36.1 &55.2 &70.9 &104.9 &105.3\\
AdaAtt\cite{lu2017knowing} (CVPR2017) &74.8&92.0&58.4&84.5&44.4&74.4&33.6&63.7&26.4&35.9&55.0&70.5&104.2&105.9\\
TD-ATT\cite{chen2017temporal} (AAAI2018) &75.7&91.3&59.1&83.6&44.1&72.6&32.4&60.9&25.9&34.2&54.7&68.9&105.9&109.0 \\
Stack-cap\cite{gu2017stack} (AAAI2018) &77.8&93.2&\textcolor{blue} {\underline{61.6}}&86.1&46.8&\textcolor{blue} {\underline{76.0}}&34.9&64.6&27.0&35.6&\textcolor{blue} {\underline{56.2}}&70.6&114.8&\textcolor{blue} {\underline{118.3}} \\
OPR-MCM\cite{zhang2019more}  (TIP2019)  &74.9&92.7&58.7&85.3&45.0&75.5&34.5&\textcolor{blue} {\underline{65.0}}&26.9&\textcolor{red} {\textbf{36.6}}&55.3&\textcolor{blue} {\underline{71.3}}&105.0&105.6\\
KMSL\cite{li2019know} (TMM2019)
&\textcolor{red} {\textbf{79.2}}&\textcolor{red} {\textbf{94.4}}&\textcolor{red} {\textbf{62.6}}&\textcolor{red} { \textbf{87.2}}&\textcolor{blue} {\underline{47.5}}&\textcolor{red} {\textbf{77.1}} &\textcolor{blue} {\underline{35.4}}&\textcolor{red} {\textbf{65.8}}&\textcolor{blue} {\underline{27.3}}&36.1&\textcolor{blue} {\underline{56.2}}&71.2&\textcolor{blue} {\underline{115.1}}&117.3 \\
TDA+GLD (Ours) &\textcolor{blue} {\underline{78.7}}&\textcolor{blue} {\underline{93.7}}&\textcolor{red} {\textbf{62.6}}&\textcolor{blue} {\underline{86.9}}&\textcolor{red} {\textbf{47.8}}&\textcolor{red} {\textbf{77.1}}
&\textcolor{red} {\textbf{35.9}}&\textcolor{red} {\textbf{65.8}}&\textcolor{red} {\textbf{27.5}}&\textcolor{blue} {\underline{36.2}}&\textcolor{red} {\textbf{56.9}}&\textcolor{red} {\textbf{71.6}}&\textcolor{red} {\textbf{116.9}}&\textcolor{red} {\textbf{119.5}} \\
\midrule
GoogleNIC  \cite{vinyals2015show} (CVPR2015) \dag &71.3&89.5&54.2&80.2&40.7&69.4&30.9&58.7&25.4&34.6 &53.0&68.2&94.3&94.6\\
SemATT \cite{you2016image} (CVPR2016) \dag &73.1&90.0&56.5&81.5&42.4&70.9&31.6&59.9&25.0&33.5&53.5&68.2&94.3&95.8\\
Rennie \cite{rennie2017self} (CVPR2017) \dag &78.1&93.1&61.9&86.0&47.0&75.9&35.2&64.5&27.0&35.5&\textcolor{blue} {\underline{56.3}}&70.7&\textcolor{blue} {\underline{114.7}}&116.7 \\
ALT-ALTM \cite{ye2018attentive} (TIP2018) \dag &74.2&92.2&57.7&84.3&44.3&74.3&34.1&63.9&27.0&37.0&55.2&71.2&105.3&105.9\\
Up-down \cite{anderson2018bottom} (CVPR2018) \dag &\textcolor{red} {\textbf{80.2}} &\textcolor{red} {\textbf{95.2}} &\textcolor{red} {\textbf{64.1}} &\textcolor{red} {\textbf{88.8}} &\textcolor{red} {\textbf{49.1}} &\textcolor{red} {\textbf{79.4}} &\textcolor{red} {\textbf{36.9}} &\textcolor{red} {\textbf{68.5}} &\textcolor{blue} {\underline{27.6}} &\textcolor{red} {\textbf{36.7}} &\textcolor{red} {\textbf{57.1}} &\textcolor{red} {\textbf{72.4}} &\textcolor{red} {\textbf{117.9}} &\textcolor{red} {\textbf{120.5}}\\
N-step SCST \cite{gao2019self} (CVPR2019) \dag & 77.6 &93.1 &61.3 &86.1 &46.5 &76.0 &34.8 &64.6 &26.9 &35.4 &56.1 &70.4 &117.4 &119.0 \\
TDA+GLD (Ours) \dag &\textcolor{blue} {\underline{79.0}}&\textcolor{blue} {\underline{94.0}}&\textcolor{blue} {\underline{63.0}}&\textcolor{blue} {\underline{87.4}}&\textcolor{blue} {\underline{48.2}}&\textcolor{blue} {\underline{77.7}}
&\textcolor{blue} {\underline{36.3}}&\textcolor{blue} {\underline{66.6}}&\textcolor{red} {\textbf{27.7}}&\textcolor{blue} {\underline{36.6}}&\textcolor{red} {\textbf{57.1}}&\textcolor{blue} {\underline{71.9}}&\textcolor{red} {\textbf{117.9}}&\textcolor{blue} {\underline{120.4}} \\
\bottomrule
\end{tabular}
\label{table:result-sever}
 \end{table*}

\subsubsection{Implementation details}
We utilize two typical and advanced methods as our reference models, i.e., Show-Tell (ST) \cite{vinyals2015show} and Top-Down Attention (TDA) \cite{anderson2018bottom}. Both models are trained with the RL-based approach in \cite{rennie2017self}. For ST, we exactly follow the details described in \cite{vinyals2015show} for implementation. For TDA, the original implementation involves an extra detector trained on the Visual Genome~\cite{krishna2017visual} dataset. For effective implementation, we remove this component and simply apply spatially adaptive max-pooling to image feature maps to obtain the final image feature. Both baseline methods adopt the encoder-decoder pipeline, and we follow existing methods \cite{rennie2017self} to use ResNet-101 \cite{he2016deep} for image encoding and an LSTM with a hidden state size of 512 for decoding captions.

We train the models by the Adam \cite{kingma2014adam} optimizer. Specifically, we follow previous works \cite{rennie2017self} to train the model using the MLE loss for the first 20 epochs and then switch to the RL loss to continue training. The batch size is set as 16, the learning rate is initialized as $5\times 10^{-4}$ and annealed by a factor of 0.8 for every 3 epochs, and the model is trained with 120 epochs in total. During inference, we use beam search with a size of 3 to decode the captions.

The reward definitions of Equations (\ref{eq:rh}) and (\ref{eq:rb}) involve computing the similarity score $s(I,c)$ of a given image and caption pair. In this work, we resort to the visual-semantic embedding method (VSE++) \cite{faghri2017vse++} to obtain the similarity score.  The hinge-based triplet ranking loss is adopted to train the VSE++ model and learn joint visual-semantic embeddings.
When training the caption generators, the parameters of VSE++ are frozen. We first extract the image feature $f_I$ using ResNet-101 \cite{he2016deep} and sentence feature $f_c$ using a gated recurrent unit (GRU) encoder \cite{cho2014learning}. Then, $f_I$ and $f_c$ are mapped into VSE++ with two linear transformations and use the cosine similarity to compute the final score.

\subsection{Comparison with State-of-the-art Methods}
The MS-COCO dataset \cite{chen2015microsoft} is the most widely used benchmark to evaluate captioning models, and most competitive works have reported their results on Karpathy's test split \cite{karpathy2015deep}. In this part, we first compare the performance of our proposed method with the following 15 state-of-the-art methods on this split.  1) GoogleNIC \cite{vinyals2015show}, which adopts a CNN-RNN architecture to directly translate image pixels to natural language descriptions. 2) MRNN \cite{mao2014deep}, which combines representation from multiple modalities. 3) SoftAtt \cite{xu2015show} and HardAtt \cite{xu2015show}, which integrate the ``soft'' deterministic attention mechanism and ``hard'' stochastic attention mechanism for learning content-related representations.
 4) SCACNN \cite{chen2016sca}, which incorporates spatial and channel-wise attentions to dynamically modulate the sentence generation context in multilayer feature maps. 5) SemATT \cite{you2016image}, which learns to selectively attend to semantic proposals and feeds them to the RNN.  6) AdaAtt \cite{lu2017knowing}, which automatically decides when to focus on the image and when to rely on the language model to generate the next word. 7) MSM \cite{yao2017boosting}, which exploits mid-level attributes as complementary information to image representation. 8) ALT-ALTM \cite{ye2018attentive}, which learns various relevant feature abstractions by attending to the high-dimensional transformation matrix from the image feature space to the context vector space. 9) OPR-MCM \cite{zhang2019more}, which adaptively re-weights the loss of different samples and uses a two-stage optimization strategy to detect more semantic concepts.
10) TD-ATT \cite{chen2017temporal}, which adopts the temporal-difference learning method to take the correlation between temporally successive actions into account for defining the reward. 11) Rennie \cite{rennie2017self}, which presents the self-critical sequence training algorithm to normalize the rewards to reduce variance in reinforcement learning. 12) Stack-cap \cite{gu2017stack}, which proposes a coarse-to-fine multistage prediction framework to produce increasingly refined image descriptions. 13) Up-down \cite{anderson2018bottom}, which combines bottom-up and top-down attention mechanisms to enable attending on semantic object cues and image features.
14) N-step SCST\cite{gao2019self}, which propose an n-step reformulated advantage function to generally increase the absolute value of the mean of reformulated advantage while lowering variance.
15) KMSL \cite{li2019know}, which takes advantage of the object entities and pairwise relationships in scene graphs for generating natural language descriptions.

\begin{table*}[t]
\centering
 \vspace{6pt}
 \caption{Ablation studies on our method using the baselines ST and TDA.}
 \begin{tabular}{c|c|c|c|c|c|c|c}
 \toprule
  \multirow{2}*{Model Variants} & \multicolumn{5}{c}{Evaluation Metrics (\%)} & \multicolumn{2}{c}{Fine-Granularity} \\ \cmidrule(lr){2-6}  \cmidrule(lr){7-8}
    &BLEU4 & ROUGEL & METEOR& SPICE& CIDEr & UniCap & AvgLen \\
  \midrule  	
  ST &32.8&54.7&25.7&19.1&103.1&2713&9.20 \\
  ST-Strengthen &32.6&54.8&25.7&19.2&102.8&2682&9.19 \\
  ST+GD &32.9&54.8&25.8&19.0&104.0&3040&9.28 \\
  ST+LD-Diff &32.9&54.8&25.8&19.0&105.9&2738&9.20 \\
  ST+LD &\textbf{33.1}&\textbf{55.0}&25.8&19.0&107.2&2765&9.22 \\
  ST+GLD &33.0&54.9&\textbf{25.9}&\textbf{19.3}&\textbf{107.7}&\textbf{3140}	&\textbf{9.29} \\
  \midrule
  TDA &36.1&57.1&27.5&21.0&117.0&3589	&9.33 \\
  TDA-Strengthen &36.2&57.0&27.4&21.2&116.8&3582	&9.29 \\
  TDA+GD &36.1&57.1&27.6&21.3&117.9&3612	&9.52	 \\
  TDA+LD-Diff &36.2&57.1&27.6&21.3&{119.3}&3513	&9.38	 \\
  TDA+LD &\textbf{36.3}&\textbf{57.2}&\textbf{27.8}&21.4&{121.0}&3448	&9.41	 \\
  TDA+GLD &36.1&57.1&\textbf{27.8}&\textbf{21.6}&\textbf{121.1}&\textbf{3797}	&\textbf{9.56}\\
 \bottomrule
 \end{tabular}
\label{table:baseline}
\end{table*}

\begin{table*}[t]
\centering
  \vspace{6pt}
 \caption{Performance of TDA+LD using different parameter settings.}
 \begin{tabular}{c|c|c|c|c|c|c|c}
 \toprule
 \multirow{2}*{Parameter Settings} & \multicolumn{5}{c}{Evaluation Metrics (\%)} & \multicolumn{2}{c}{Fine-Granularity} \\ \cmidrule(lr){2-6}  \cmidrule(lr){7-8}
  \  &BLEU4 & ROUGEL & METEOR& SPICE& CIDEr & UniCap & AvgLen \\
  \midrule	
  TDA+LD ($\lambda$=0; $\eta$=0) &36.1&57.1&27.5&21.0&117.0&3589&9.33 \\
  TDA+LD ($\lambda$=2; $\eta$=0) &35.1&56.3&27.3&20.6&119.7&3165&9.29 \\
  TDA+LD ($\lambda$=5; $\eta$=0) &36.0&57.0&\textbf{27.8}&21.2&120.5&3340&9.38 \\
  TDA+LD ($\lambda$=8; $\eta$=0) &35.7&56.8&27.6&21.0&120.2&3276&9.34 \\ \hline
  TDA+LD ($\lambda$=5; $\eta$=0) &36.0&57.0&\textbf{27.8}&21.2&120.5&3340&9.38 \\
  TDA+LD ($\lambda$=5; $\eta$=1) &\textbf{36.3}&\textbf{57.2}&\textbf{27.8}&\textbf{21.4}&\textbf{121.0}&\textbf{3448}	 &\textbf{9.41}	 \\
  TDA+LD ($\lambda$=5; $\eta$=2) &36.1&57.0&\textbf{27.8}&21.2&120.7&3394&9.38 \\
 \bottomrule
 \end{tabular}
\label{table:Parameters}
\end{table*}

The performance results on Karpathy's test split are shown in Table \ref{table:result}.
As can be observed, the previous leading methods are Stack-cap \cite{gu2017stack} and Up-down \cite{anderson2018bottom}, which obtain the CIDEr score of 120.4\% and 120.1\%, respectively.
Our approach outperforms these competitors in terms of ROUGEL, METEOR, SPICE and CIDEr.
Furthermore, our approach improves the CIDEr score to 121.1\%.
For a more comprehensive comparison, we also submit our result to the online MS-COCO test server for evaluation, and we summarize the results of our method and those of the published leading competitors in Table \ref{table:result-sever}.
When using a single model, our method achieves the best performance among the competitors on most evaluation metrics, as exhibited in the upper part of Table \ref{table:result-sever}. Some methods also report the results of ensembling several models \cite{anderson2018bottom,rennie2017self}. By simply ensembling four models, our method achieves competitive performance with the existing state-of-the-art ensemble method \cite{anderson2018bottom}, ranking first or second among the competitors on all the metrics.
Notably, \cite{anderson2018bottom} achieves better results than our method on some evaluation metrics because it additionally utilizes an object-detection-based attention mechanism.
Work \cite{anderson2018bottom} detects objects from the image via a pre-trained detector and employs an attentional mechanism to infer the useful semantic regions. It can locate the semantic regions more accurately and thus facilitate image captioning performance. However, this method relies on the detector that requires additional annotations and is more complicated as it needs to execute the additional detector during inference. In contrast, our method directly applies attentional mechanism at the final convolutional feature maps and introduces no additional annotation and computing overhead.


\subsection{Ablation Study}
The proposed global-local discriminative (GLD) objective is formulated based on existing reference models and consists of two components, i.e., a global discriminative (GD) constraint and a local discriminative (LD) constraint. In this section, we first conduct experiments to compare with the baseline models to demonstrate the overall effectiveness of the proposed GLD objective and then perform further analysis to assess the contribution of each component.
\subsubsection{Overall Contribution of the Global-Local Discriminative (GLD) Objective} 
\begin{figure}[t]
\centering
\subfigure[]{
\includegraphics[width=0.46\linewidth]{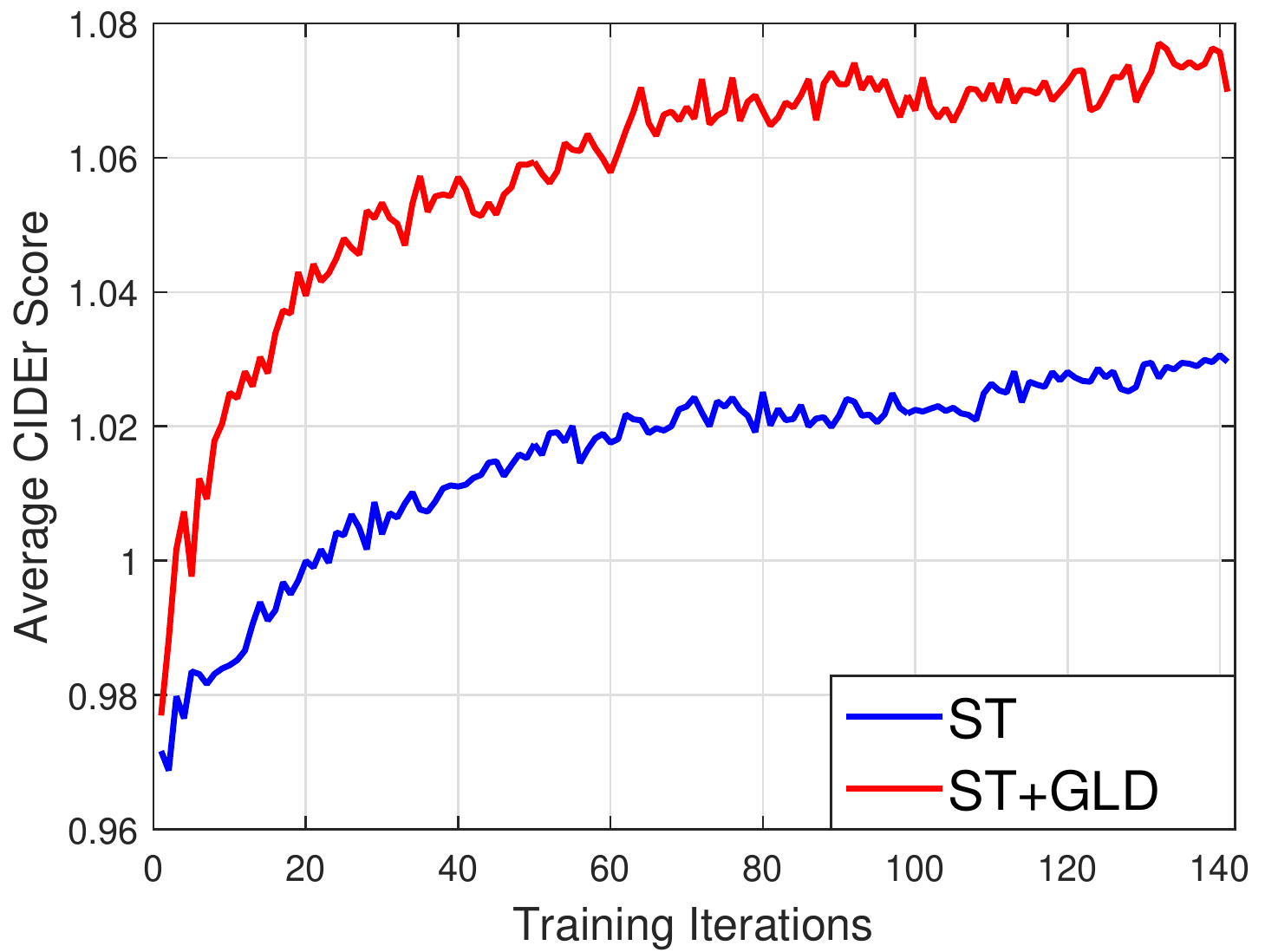}}
\subfigure[]{
\includegraphics[width=0.46\linewidth]{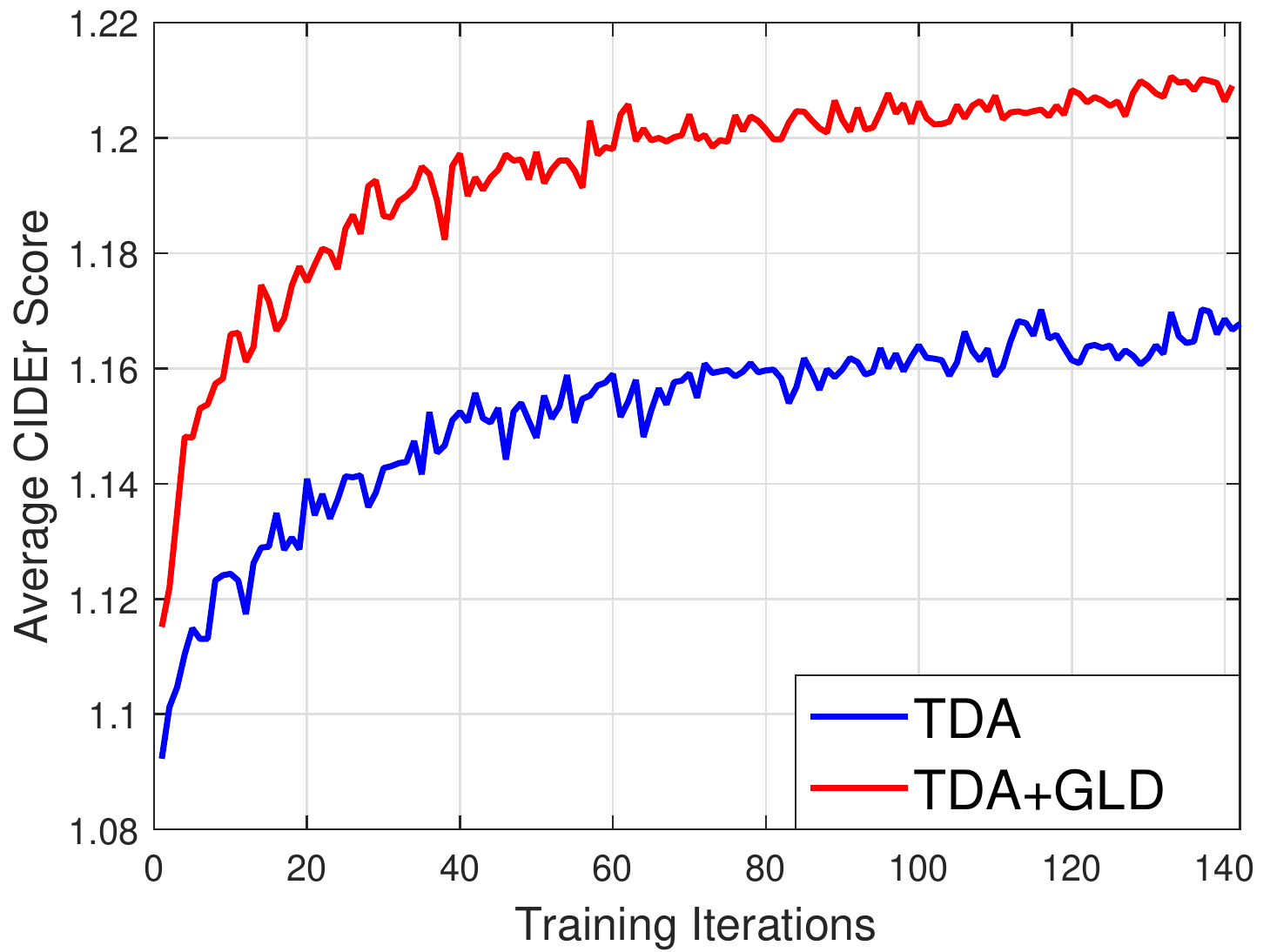}}
\caption{Average CIDEr score curve of our method and the baseline for the (a) ST and (b) TDA reference models.}
\label{fig:CIDER}
\end{figure}

We perform an in-depth comparative analysis with the reference models to demonstrate the effectiveness of our GLD objective. Specifically, we compare with ST \cite{vinyals2015show} and TDA \cite{anderson2018bottom}, the two baseline reference models that we use, and present the comparison results on the MS-COCO dataset in Table \ref{table:baseline}.
As can be observed, our method exhibits notable improvements in all metrics, e.g., improving the CIDEr score by 4.6\% when using the ST baseline and by 4.1\% when using the TDA baseline. To better illustrate the performance comparison with the baseline methods, we further show the curves of CIDEr score v.s. training iteration in Figure \ref{fig:CIDER}.
Because the result of the MS-COCO test server is submitted online and the number of submissions allowed is limited, thus we cannot obtain the result of each training iteration.
Hence Figure \ref{fig:CIDER} reflects the result of the Karpathy's test split, which is a common test set for this task.
We find that our method converges faster and consistently outperforms the baseline method by a sizable margin during the entire training process.

To evaluate the fine-granularity of the generated captions, two quantitative metrics (i.e., UniCap and AvgLen) are introduced. The UniCap metric denotes the number of unique captions generated by a tested method on the test set. The larger the UniCap is, the more powerful the model is to generate fine-grained captions that discriminate among similar images. A sufficiently good caption model should generate a distinct caption for each different image.
The AvgLen metric denotes the average length of all the captions generated by a tested method on the test set.  It can reflect the fine-granularity of captions to some extent since a more fine-grained caption is generally longer.
From the results summarized in Table \ref{table:baseline}, it can be inferred that our method encourages generating more discriminative and fine-grained descriptions than the baselines. Specifically, upon the 5,000 test images, our method improves the UniCap from 2,713 to 3,140 for the ST baseline and from 3,589 to 3,797 for the TDA baseline. It verifies that our method has a significant effect on discriminative caption generation. With respect to the average length of captions, our method achieves an increase of 0.09 for ST and 0.23 for TDA, which demonstrates the effectiveness of our method from another perspective.
These results also show that our proposed objective can achieve a consistent improvement on different baselines, indicating our method's good adaptability to various caption models.

\begin{figure}[!t]
\centering
\includegraphics[width=0.95\linewidth]{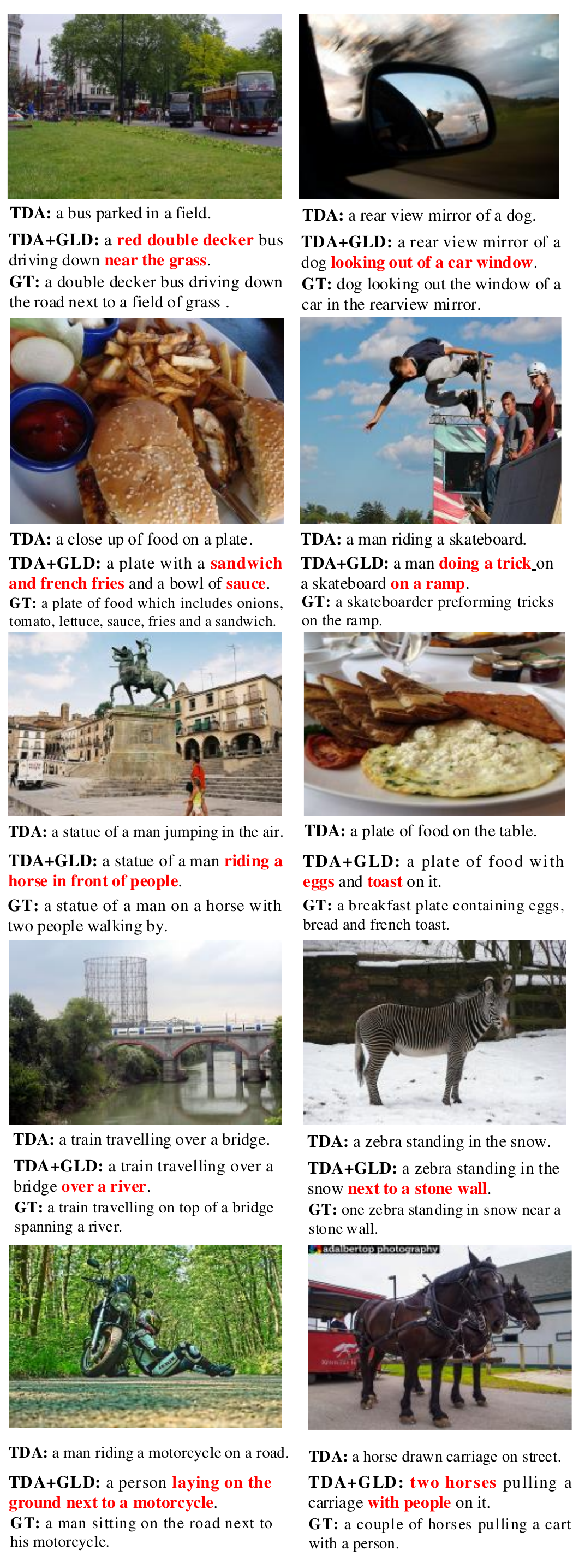}
\caption{Captions generated by the TDA baseline and our TDA+GLD.}
\label{fig:vis1}
\end{figure}

\begin{figure}[!t]
\centering
\includegraphics[width=0.95\linewidth]{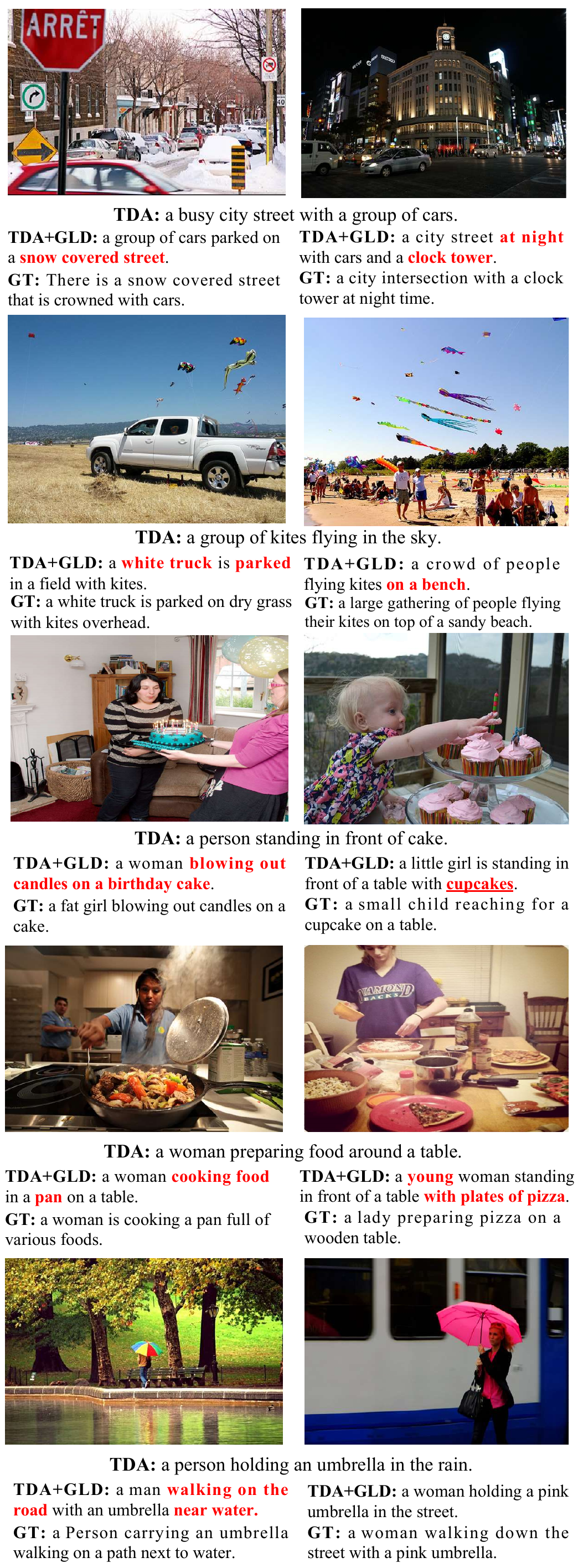}
\caption{Captions generated by the TDA baseline and our TDA+GLD.}
\label{fig:vis2}
\end{figure}

We further visualize some representative results to provide a more direct qualitative comparison. Figure \ref{fig:vis1} exhibits a few test images, corresponding ground-truth (GT) captions, and the captions generated by the baseline TDA and our method (TDA+GLD). It can be seen that our method can generate more fine-grained and detailed captions that better describe the given images.
 Taking the image in the upper left of Figure \ref{fig:vis1} as an example, the TDA baseline ignores the content of the grass, but our method can well describe this detail. Moreover, our method generates the more fine-grained and accurate phrase ``red double decker bus'', while TDA just predicts a common word ``bus''.  A similar phenomenon can also be observed in other examples.

Moreover, when the baseline method tends to generate the same caption for similar images, our method can generate discriminative captions. Some examples are presented in Figure \ref{fig:vis2}. For the two images in the second row of Figure \ref{fig:vis2}, the TDA baseline merely describes the shared content ``flying kites in the sky'' and generates a sentence ``a group of kites flying in the sky'' for both images. In contrast, except for describing the shared content, our method can further capture ``parked truck'' for the first image and ``people crowd and bench'' for the second image.
It shows that our method can capture more details and thus generate more fine-grained and distinguishable captions.


\subsubsection{Contribution of Global Discriminative (GD) Constraint}
We merely incorporate the GD constraint in the reference model to evaluate its contribution. As shown in Table \ref{table:baseline}, it can improve most of the evaluation metrics. For the more comprehensive metric (CIDEr) , the score is increased by 0.5\% with the ST baseline and by 0.9\% with the TDA baseline. Furthermore, the GD constraint can significantly increase the fine-granularity of generated captions. Specifically, it increases the number of unique sentences from 2,713 to 3,040 for the ST baseline and from 3,589 to 3,612 for the TDA baseline.

\begin{figure}[t]
\centering
\subfigure[]{
\includegraphics[width=0.46\linewidth]{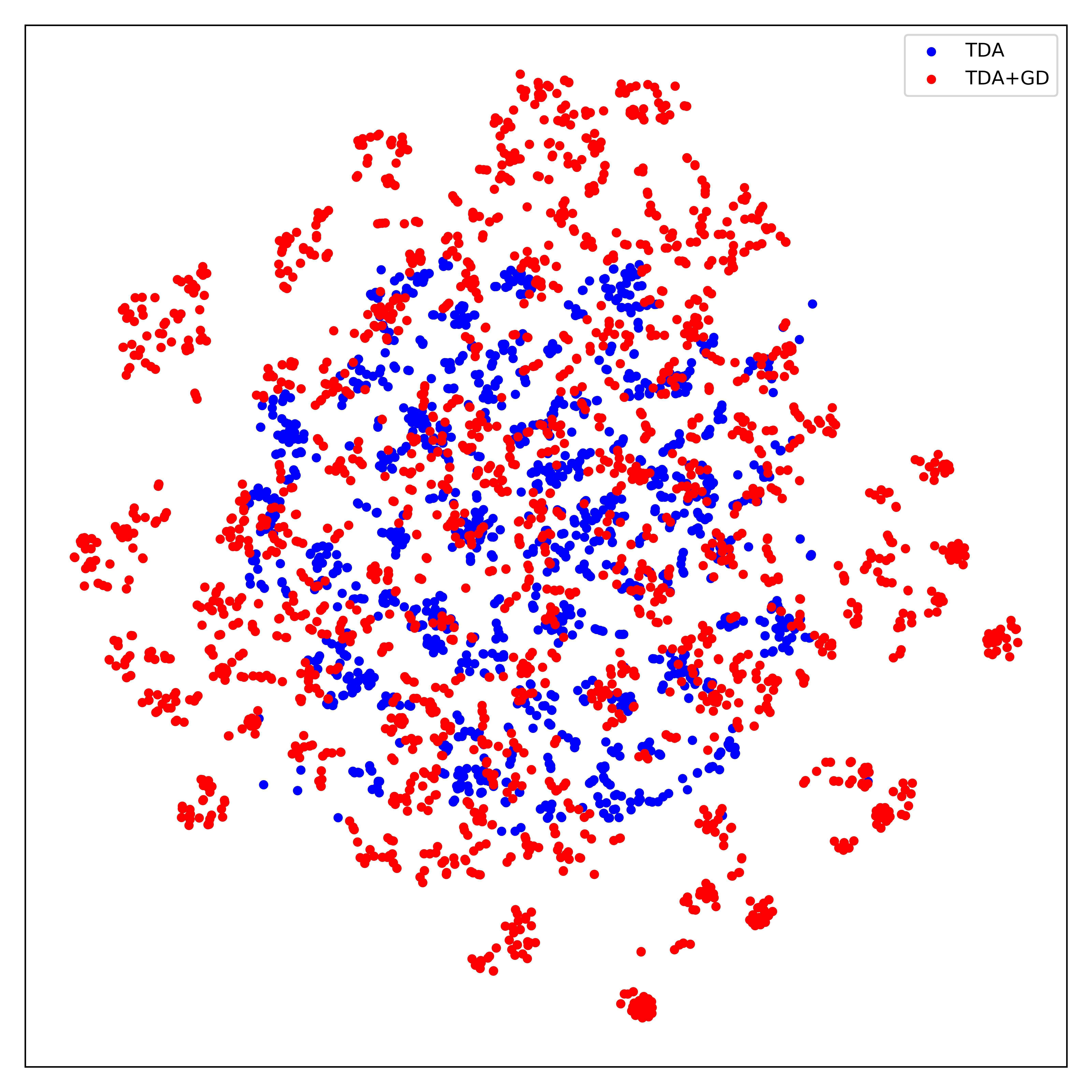}}
\subfigure[]{
\includegraphics[width=0.46\linewidth]{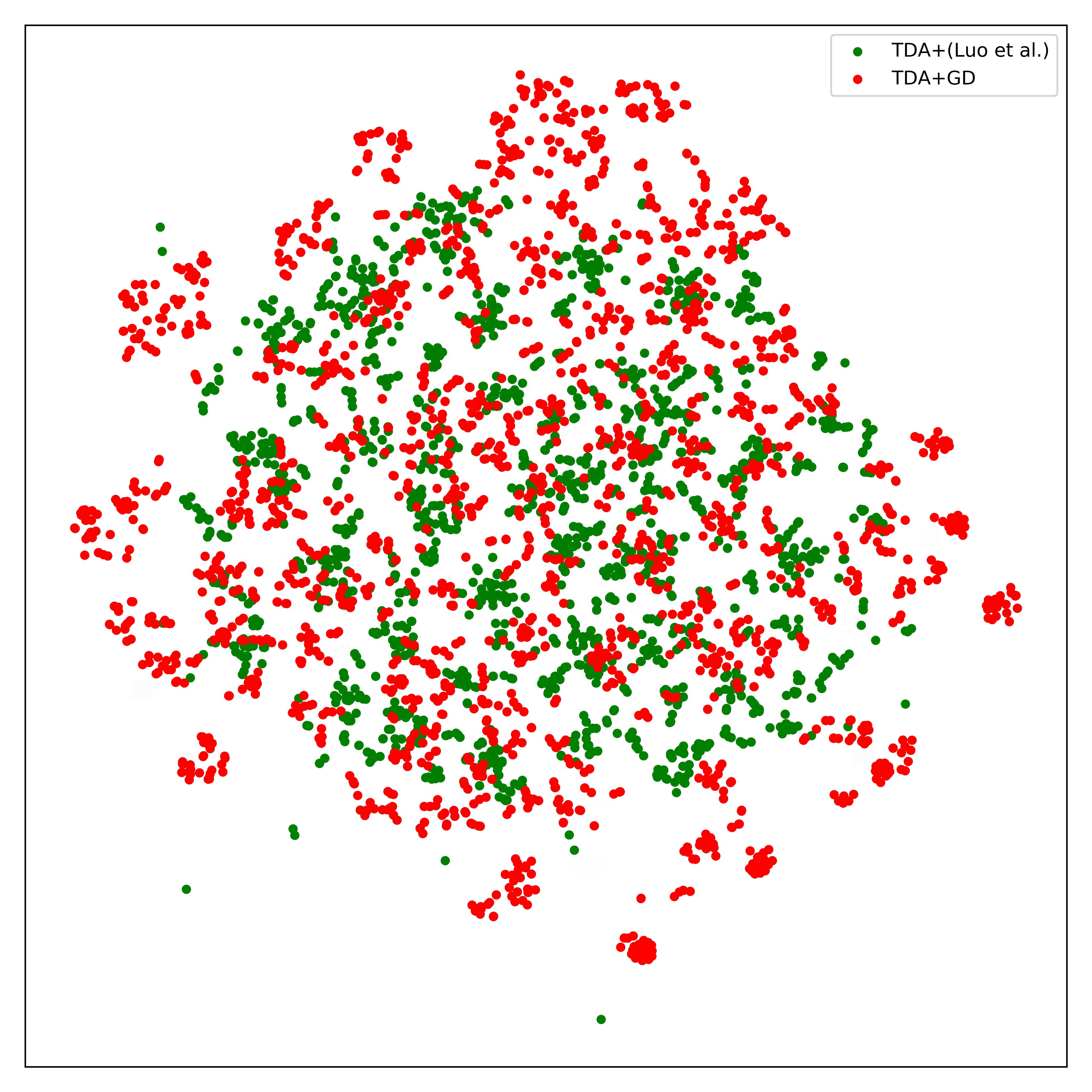}}
\caption{Embedding of the caption space visualized using t-SNE: (a) TDA and TDA+GD; (b) TDA+(Luo et al.) and TDA+GD.}
\label{fig:tsne}
\end{figure}

To further demonstrate the effectiveness of the global discriminative constraint, we use the t-distributed stochastic neighbor embedding (t-SNE) \cite{maaten2008visualizing} visualization technique to analyze the discriminability of captions in Figure \ref{fig:tsne}. Specifically, we use the proposed model to generate captions for 5,000 images in the Karpathy test set, and apply the GRU encoder from visual-semantic embedding \cite{faghri2017vse++} to extract a 1,024-dimension representation vector for each generated caption. Then, the t-SNE \cite{maaten2008visualizing} method is applied to reduce the vector dimension and visualize the caption representation distribution in Figure \ref{fig:tsne}.
Intuitively, we can observe that the caption representation cluster produced by TDA+GD is more dispersed than that of TDA and TDA+(Luo et al.) \cite{luo2018discriminability}. It shows that the captions generated by TDA and TDA+(Luo et al.) are of less variance and more similarity, and indicates that the captions provided by our TDA+GD are more discriminative. In Section \ref{sec:Self-Retrieval}, further quantitative evaluation will be provided to demonstrate the effectiveness of our GD constraint.

\subsubsection{Contribution of Local Discriminative (LD) Constraint}
We evaluate the contribution of the LD constraint by merely incorporating it in the reference model. As shown in Table \ref{table:baseline}, it leads to a clear performance improvement, e.g., obtaining an obvious CIDEr increase of 4.1\% and 4.0\% over the ST and TDA reference models, respectively.
As described above, our LD constraint works by introducing a word-level reward assignment mechanism, which assigns higher rewards to the more fine-grained and content-sensitive words/phrases that describe the visual details of given images. Below we perform further investigation to analyze the formulation of our LD constraint.

First, our LD constraint strengthens the TF-IDF weights of the less frequent but informative words. A question that may arise is whether the benefit of our LD constraint can come from simply strengthening TF-IDF weights in CIDEr \cite{vedantam2015cider}. Thus, we design a new baseline with such a TF-IDF weight adjustment. Specifically, the logarithmic base $e$ in CIDEr is replaced by 2 to strengthen TF-IDF weights in CIDEr, and then the original caption-level reward is still used for each word on the ST/TDA baselines. The new baselines are denoted as ST-Strengthen/TDA-Strengthen.
As exhibited in Table \ref{table:baseline}, the performance of ST-Strengthen/TDA-Strengthen is just comparable with the original ST/TDA and even worsens in some metrics. It shows that the key factor is not that the IDF weighting in CIDEr is too weak, and that only strengthening TF-IDF weights in CIDEr cannot contribute to our LD constraint.

Second, we propose the idea that provides each word a word-level CIDEr reward in our LD constraint. To better verify the effectiveness of this idea, we design a simpler variant of our LD constraint, termed as ST+LD-Diff/TDA+LD-Diff. Specifically,
we calculate the difference between the previous and the current caption scores when each new word is generated and appended, and this difference is used as a word-level reward for the newly appended word. Consequently, Equation (\ref{eq:LD}) becomes $R_{\mathrm{LD2}}(w^s_t)= R_{\mathrm{C}}(c_{1:t}) - R_{\mathrm{C}}(c_{1:t-1}) + R_{\mathrm{C}}(\tilde{c})$, where $c_{1:t}$ denotes the sentence fragment from 1 to $t$ in the time series.
From Table \ref{table:baseline}, we can see that ST+LD-Diff/TDA+LD-Diff performs better than ST/TDA, achieving a CIDEr increase of 2.8\% and 2.3\%, respectively. It demonstrates that the idea of word-level rewards does play a significant role in our LD constraint. It can also be easily observed that, by using our LD constraint, ST+LD/TDA+LD obtain still better results than ST+LD-Diff/TDA+LD-Diff, further increasing the CIDEr by 1.3\% and 1.7\%, respectively. The reason is that our LD constraint also strengthens the weights of less frequent but informative words and takes phrases into account, while the difference-based word-level reward only considers the effect of single words.

\begin{figure}[t]
\centering
\includegraphics[width=0.95\linewidth]{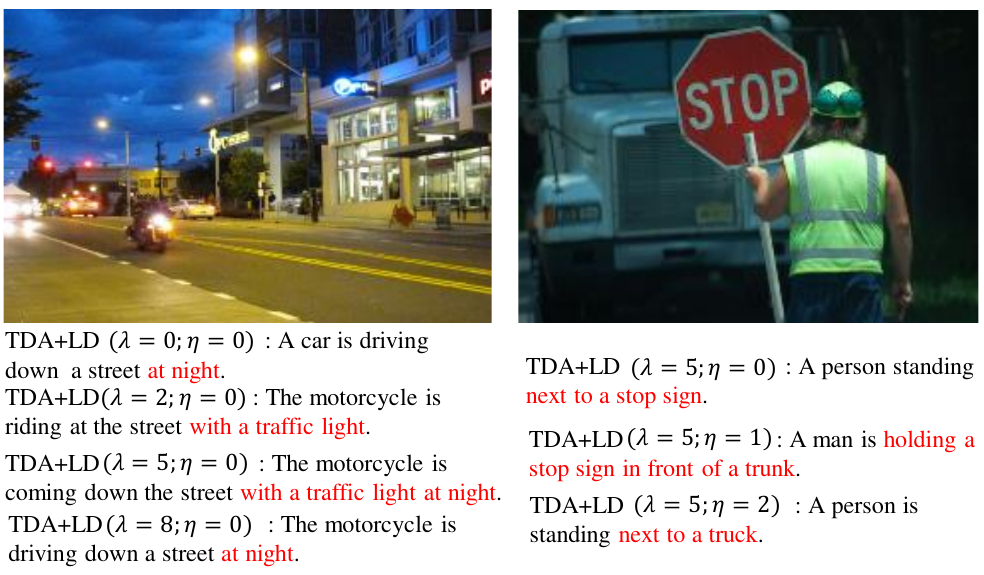}
\caption{Captions generated by TDA baseline with different thresholds.}
\label{fig:threshold}
\end{figure}

Third, our LD constraint adopts two thresholds to help select the fine-grained words, i.e., $\eta$ and $\lambda$ in Equation (\ref{eq:LD}). Therefore, we conduct experiments to analyze the influence of different threshold settings on performance. We first fix $\eta=0$ and increase $\lambda$ from 2 to 8. As shown in Table \ref{table:Parameters}, we find that the performance first increases when increasing $\lambda$ from 2 to 5, but the performance becomes worse with a further increase of $\lambda$. Thus, setting $\lambda=5$ can roughly achieve the best performance. Then, we fix $\lambda$ as 5 and increase $\eta$ from 0 to 2. The results are also presented in Table \ref{table:Parameters}. A similar performance change is observed, and we find that setting $\eta=1$ roughly achieves the best performance.
Based on the above analysis, we set $\lambda=5$ and $\eta=1$ in all the experiments.
To further qualitatively validate the performance of $\lambda$ and $\eta$, we randomly choose two test images and show the results with different threshold settings in Figure \ref{fig:threshold}. We can find that setting a small or large value for $\lambda$ and $\eta$ will tend to generate less-informative captions, which is consistent with the quantitative analysis.
The possible reason is summarized as follows. In the training stage, the word-level rewards will be increased for almost all the words when $\lambda$ and $\eta$ take small values, and some fine-grained phrases will be ignored when $\lambda$ and $\eta$ take large values.
Both cases will not take care of the fine-grained phrases well.

\subsection{Evaluation on Self-Retrieval} \label{sec:Self-Retrieval}

In this subsection, we follow previous work \cite{dai2017contrastive} to conduct self-retrieval experiments to assess the discriminability of the proposed approach. Specifically, we first randomly select 5,000 images $\{I_1, I_2, \dots, I_{5000}\}$ from the MS-COCO test set and use the caption model to generate the corresponding 5,000 sentences $\{c_1, c_2, \dots, c_{5000}\}$. Then, for each sentence $c_i$, we use it as a query and compute the probabilities conditioned on each image, i.e., $\{p(c_i|I_1), p(c_i|I_2), \dots, p(c_i|I_{5000})\}$.
If the conditional probability $p(c_i|I_i)$ is within the top-K highest probabilities, the image $I_i$ is considered to be a top-K recalled image. The Recall@K (R@K) metric is used to measure the model's discriminability, and it is defined as the fraction of the top-K recalled images relative to all the 5,000 images.
A higher Recall@K indicates that more images are easily retrieved by their corresponding generated sentences, and thus, the caption model captures the distinctiveness of images better.

\begin{figure}[!t]
\centering
\subfigure[]{
\includegraphics[width=0.46\linewidth]{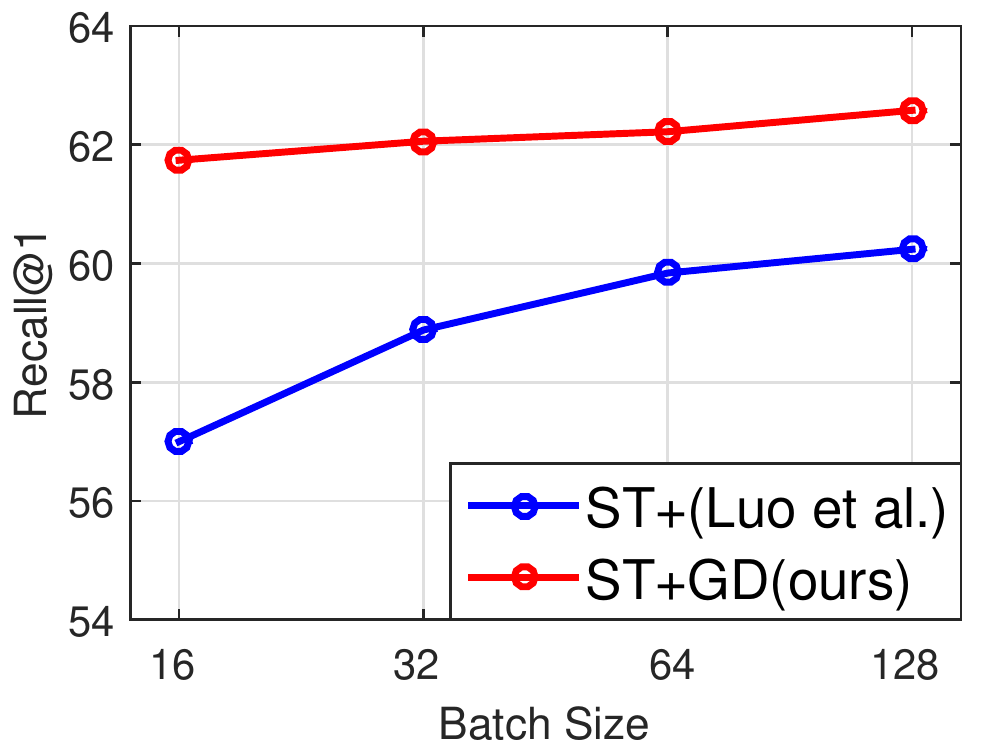}}
\subfigure[]{
\includegraphics[width=0.46\linewidth]{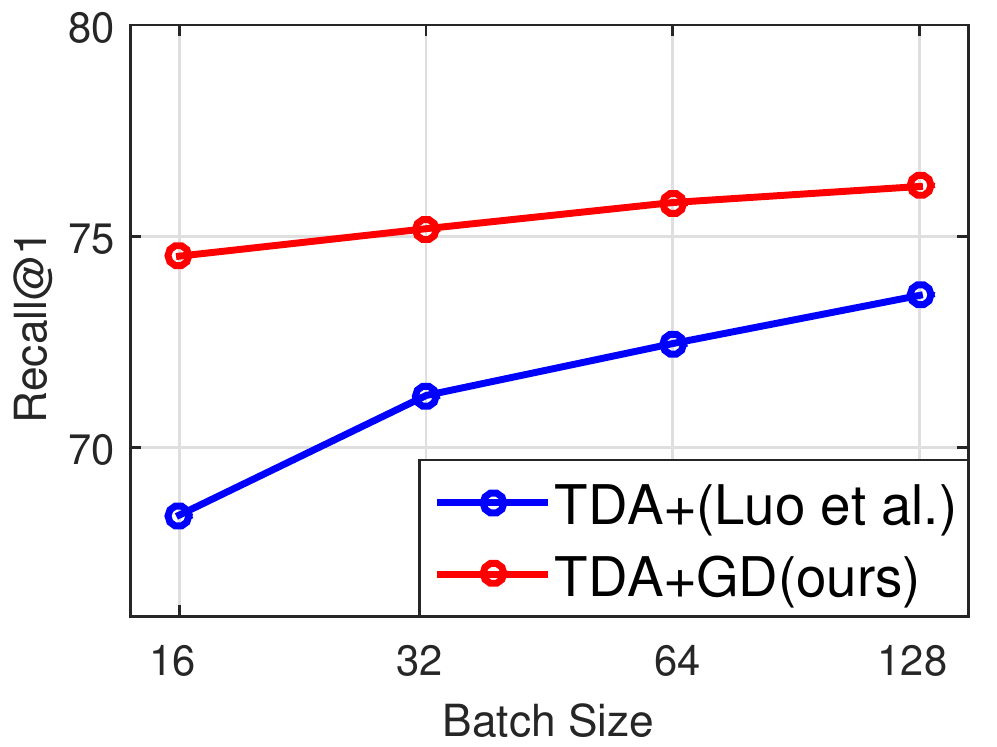}}
\caption{Recall@1 comparison between our GD method and Luo et al. \cite{luo2018discriminability} with different batch sizes using the (a) ST and (b) TDA baselines.}
\label{fig:r1}
\end{figure}

\begin{figure}[!t]
\centering
\subfigure[]{
\includegraphics[width=0.46\linewidth]{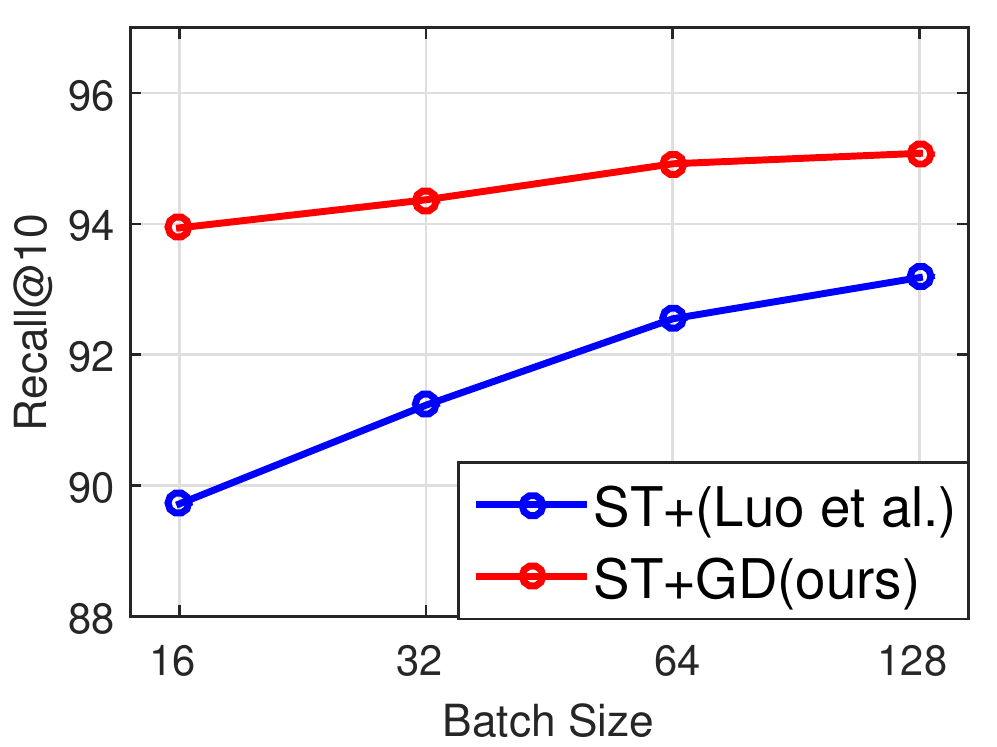}}
\subfigure[]{
\includegraphics[width=0.46\linewidth]{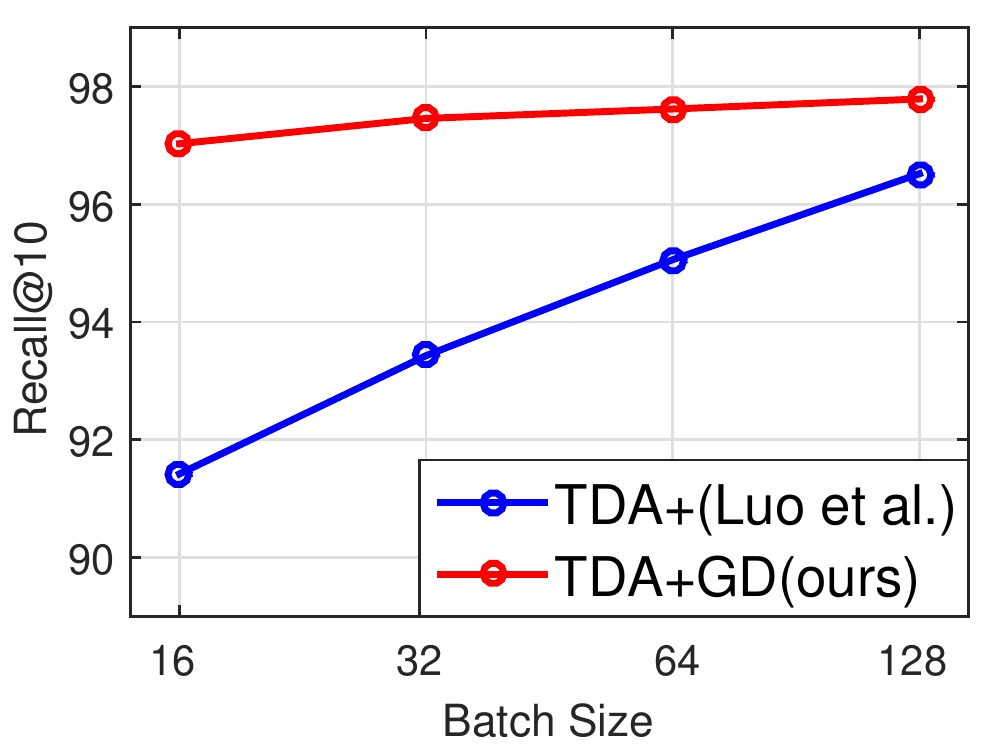}}
\caption{Recall@10 comparison between our GD method and Luo et al. \cite{luo2018discriminability} with different batch sizes using the (a) ST and (b) TDA baselines.}
\label{fig:r10}
\end{figure}

The evaluation results are reported in Table \ref{table:retrieval}. Since previous work (Luo et al.) \cite{luo2018discriminability} also employed a ranking loss to improve the discriminability, we implement their loss on the ST and TDA baselines for comparison. As shown in Table \ref{table:retrieval}, by introducing the ranking loss, ST/TDA+\cite{luo2018discriminability} outperform the baseline methods and achieve an impressive retrieval performance. The better-performing model TDA+\cite{luo2018discriminability} obtains the Recall@1, Recall@5, and Recall@10 of 73.6\%, 93.04\%, and 96.02\%, respectively.
Compared with \cite{luo2018discriminability} that only uses a ranking loss on the minibatch, our GD constraint additionally contains a ranking loss defined on the entire training set. It can be observed that our GD constraint achieves better results than \cite{luo2018discriminability} on both ST and TDA baselines. Moreover, we note that \cite{luo2018discriminability} requires setting a large batch size during training to ensure performance.
As exhibited in Figures \ref{fig:r1} and \ref{fig:r10}, the Recall@1 and Recall@10 performance are plotted with respect to different batch sizes. ST/TDA+(Luo et al.) \cite{luo2018discriminability} suffer a significant performance drop when the batch size is decreased. In contrast, our GD method achieves more steady results with different batch sizes.

\begin{table}[!t]
\centering
  \vspace{6pt}
\caption{Performance comparison on self-retrieval.}
\newcommand {\tabincell}[2]{\begin{tabular}{@{}#1@{}}#2\end{tabular}}

 \begin{tabular}{c|c|c|c}
 \toprule
 \multirow{2}*{Model} & \multicolumn{3}{c}{Performance (\%)} \\ \cline{2-4}
 \ &R@1 & R@5 & R@10 \\
 \hline
\ ST+\cite{luo2018discriminability} &60.24&86.33&93.18  \\
\hline
\ ST &50.84&78.54&87.58\\
\ ST+LD &53.60&82.24&90.38 \\
\ ST+GD &61.74&87.12&93.94\\
\ ST+GLD &\textbf{62.08}&\textbf{88.24}&\textbf{94.36}\\
 \midrule
  TDA+\cite{luo2018discriminability} &73.60&93.04&96.52 \\
  \hline
\ TDA &66.40&88.46&94.26 \\
\ TDA+LD &68.82&90.90&95.80 \\
\ TDA+GD &74.53&93.67&97.03 \\
\ TDA+GLD &\textbf{76.24}&\textbf{94.50}&\textbf{97.90}\\
 \bottomrule
 \end{tabular}
\label{table:retrieval}
\end{table}

\begin{table*}[t]
	\centering
	\caption{ Per-Batch Training Time (s) for some baselines. - denotes that this model does not require this step.}
	\newcommand {\tabincell}[2]{\begin{tabular}{@{}#1@{}}#2\end{tabular}}
	\begin{tabular}{c|c|c|c}
		\toprule
		\ Model & Finding Similar Images & Reward Reweighting & Backpropagation \\
		\hline
		TDA+\cite{luo2018discriminability} & 0.106 &-& 0.368 \\
		\hline
		\ TDA &-&-& 0.323  \\
		\ TDA+LD &-& 0.177 & 0.362  \\
		\ TDA+GD &0.139&-&0.372 \\
		\ TDA+GLD &0.139&0.177&0.385 \\
		\bottomrule
	\end{tabular}
	\label{table:cost}
\end{table*}

We further compare the variants of our method, which merely uses the GD or LD constraint. As shown in Table \ref{table:retrieval}, both the GD and LD constraints can achieve a notable performance gain on the baseline methods and thus help improve the model's discriminability. It is also easily observed that the GD constraint performs better than the LD constraint. For example, ST+GD increases from ST by 10.90\%, 8.58\% and 6.36\% on Recall@1, Recall@5, and Recall@10, respectively, compared to 2.76\%, 3.70\% and 2.80\% for ST+LD. The reason is that our GD and LD constraints focus on different perspectives. The GD constraint directly models the contrast among similar images and guides the caption model to generate sentences that describe the major discriminative image content, while the LD constraint drives the caption model to add more fine-grained phrases that are sensitive to content details and improve the discriminability indirectly.  By combining both, our overall GLD objective can further improve the discriminative performance, as demonstrated in Table \ref{table:retrieval}.

\subsection{Computational Costs}
The training procedure for TDA+GLD consists of three steps:
1) finding similar images to compute ranking loss,
2) reweighting reward to obtain the word-level reward, and 3) backpropagation to update parameters.
To evaluate the computational cost of each step, we report the per-batch training times of TDA+\cite{luo2018discriminability}, TDA, TDA+LD, TDA+GD, and TDA+GLD in Table \ref{table:cost}. All algorithms are implemented in PyTorch and run on an NVIDIA TITAN card with a minibatch size of 16. As outlined in Table \ref{table:cost}, we have the following observations:
(i) TDA+\cite{luo2018discriminability}, TDA+GD, and TDA+GLD indeed consume some computational costs in finding similar images. However, the cost can be negligible as the similarity matrix of images can be precomputed.
(ii) The reweighting reward consumes more computational resources. But the TF-IDF weights of the n-grams can also be precomputed to save time.
(iii) TDA+GLD helps to provide more accurate and discriminative captions, and it can achieve a better tradeoff between computational cost and caption quality.

\subsection{Discussion and Limitations}
\begin{figure}[!t]
	\centering
	\includegraphics[width=0.95\linewidth]{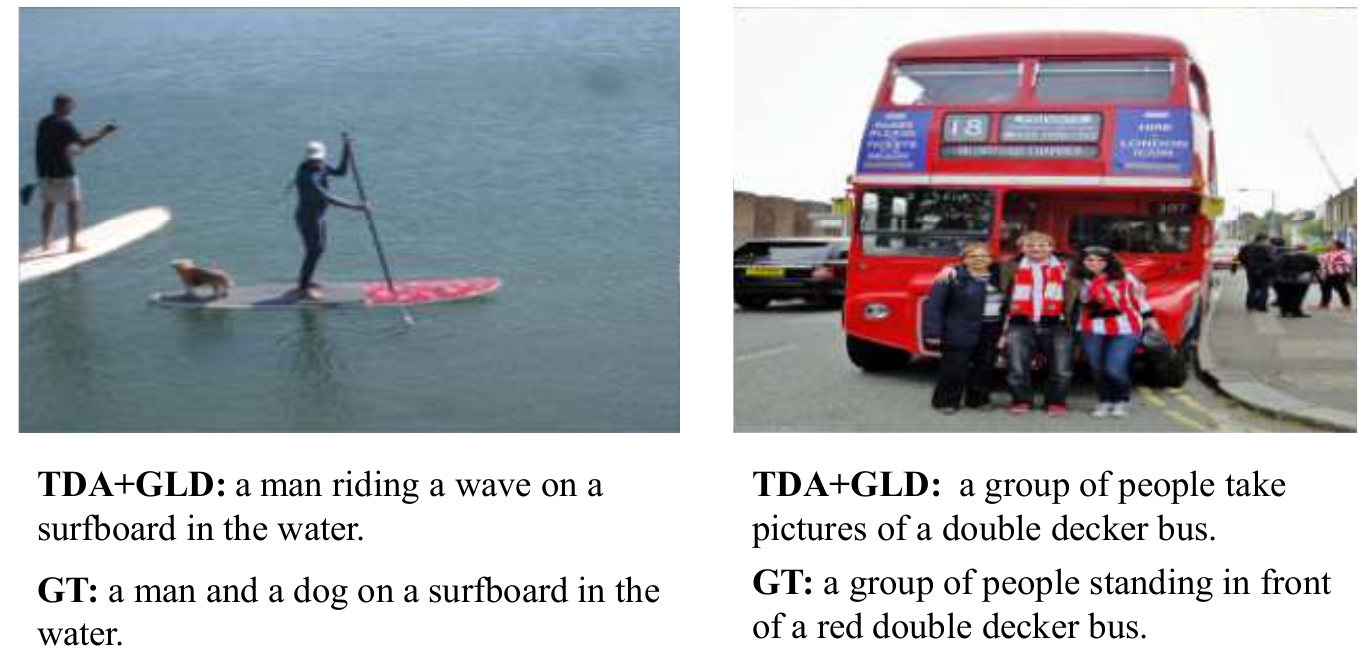}
	\caption{Two inaccurate captions generated by the TDA+GLD.}
	\label{fig:vis3}
\end{figure}
	
In this work, we address the fine-grained image captioning task. But, in fact, there is no unified evaluation metric for this task. So we introduce some metrics to measure ``fine-grained captioning''. First, inspired by \cite{luo2018discriminability}, UniCap and AvgLen are introduced to reveal the fine-granularity of the generated captions to some extent. Second, we follow some works from natural language processing \cite{maaten2008visualizing} to visualize the caption representation distribution via the t-SNE technology, where a more dispersed cluster indicates the generated captions are more discriminative. Third, we follow previous work \cite{dai2017contrastive} to perform self-retrieval experiments that quantitatively measure the discriminability of the generated captions. Maybe the proposed metrics cannot fully measure the fine-grained captioning, but they have a great deal of relevance and reliability. Our extensive qualitative and quantitative analysis can demonstrate fine-grained captioning to a large extent. We believe that our work plays a very important role in the early exploration of the task.	

There are some limitations in our approach for addressing fine-grained image captioning. We find that our method tends to generate sentences that are more discriminative than others, and thus some generated sentences do not well match the ground truth description of the images. In Figure \ref{fig:vis3}, we present some examples that our method generates unsatisfying captions.  One possibility for this phenomenon is that the global discriminative (GD) constraint pulls all the generated sentences away from each other, and thus it tends to generate captions describing discriminative contents. In fact, discriminability varies for different images. Thus, we will explore an adaptive GD constraint that gives different balance factors to the GD constraint for different images to avoid this phenomenon.

\section{Conclusion}
To generate fine-grained and discriminative captions, we propose a global-local discriminative objective, which is formulated as two constraints based on a reference model. Specifically, the global discriminative constraint is utilized to pull the generated caption to better describe the distinctiveness of the corresponding image, thus improving the discriminability, and the local discriminative constraint is designed to focus more on the less frequent words and thus enable describing more detailed and fine-grained contents of the input image. Extensive experimental evaluation and comparison with existing leading methods on the MS-COCO dataset demonstrate the effectiveness of the proposed method. However, the proposed global-local discriminative objective has limitations that should be addressed in the future. For example, the thresholds of the local discriminative constraint are set empirically and establishing how to adaptively adjust these thresholds remains to be studied.
Moreover, our discriminative objective helps with informative but less frequent words, but it will also suppress informative and highly frequent words. Therefore, a more general and adaptive discriminative objective needs to be investigated.


%



\ifCLASSOPTIONcaptionsoff
  \newpage
\fi

\bibliographystyle{IEEEtran}
\bibliography{bib}

%

\begin{IEEEbiography}[{\includegraphics[width=1in,height=1.25in,clip,keepaspectratio]{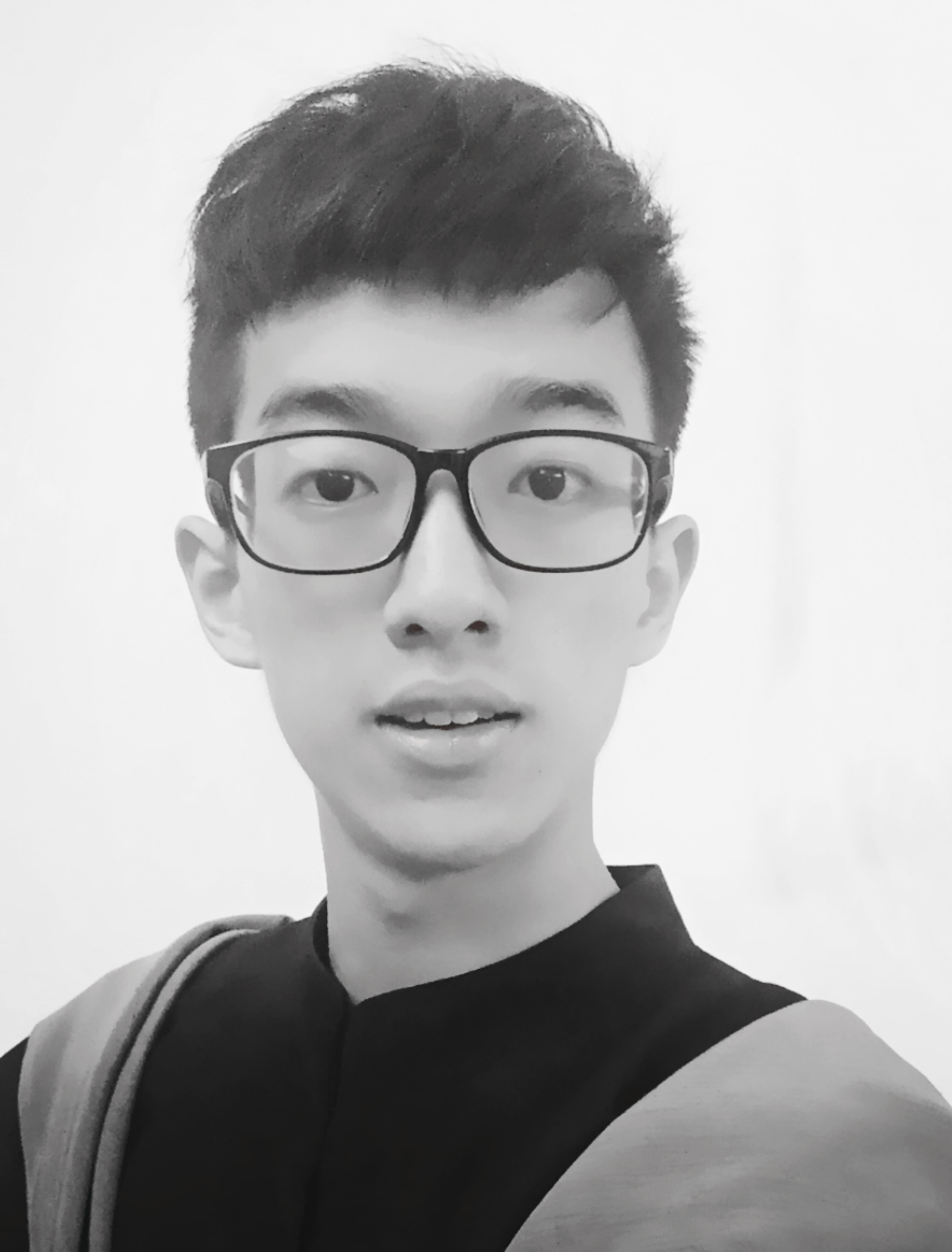}}]{Jie Wu}
received his B.E. degree from the College of Engineering, Shantou University, Shantou, China, in 2017. He is currently pursuing his Master's Degree in the School of Electronics and Information Engineering, Sun Yat-sen University, China. His current research interests include computer vision and natural language processing.
\end{IEEEbiography}
\vspace{-20pt}

\begin{IEEEbiography}[{\includegraphics[width=1in,height=1.25in,clip,keepaspectratio]{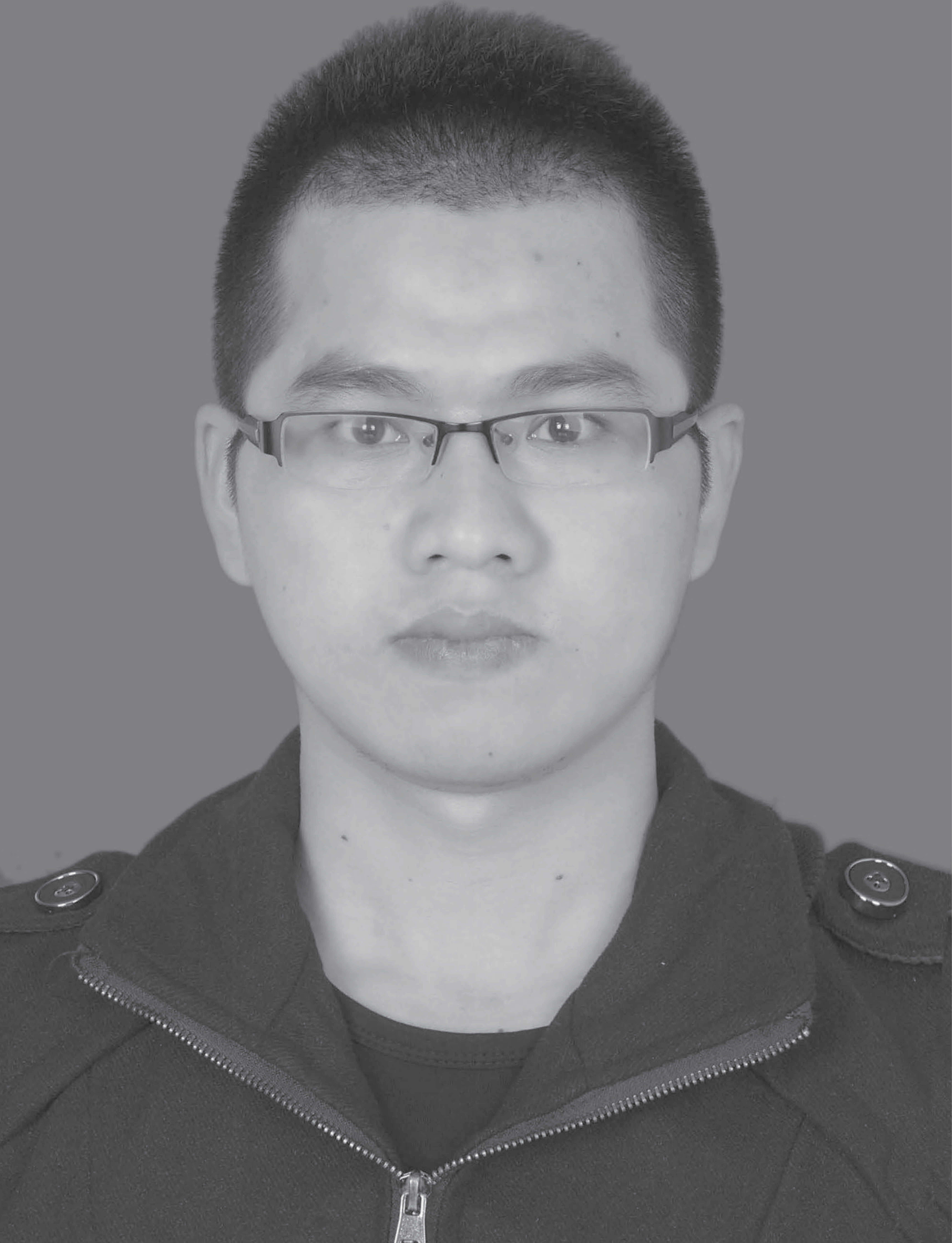}}]{Tianshui Chen}
received a Ph.D. degree in computer science at the School of Data and Computer Science Sun Yat-sen University, Guangzhou, China, in 2018. Before that, he received a B.E. degree from the School of Information and Science Technology. He is currently an Associate Research Director at DMAI Co., Ltd. His current research interests include computer vision and machine learning. He has authored and co-authored approximately 20 papers published in top-tier academic journals and conferences. He has served as a reviewer for numerous academic journals and conferences, including TPAMI, TIP, TMM, TNNLS, CVPR, ICCV, ECCV, AAAI and IJCAI. He was the recipient of the Best Paper Diamond Award at IEEE ICME 2017.
\end{IEEEbiography}
\vspace{-20pt}
\begin{IEEEbiography}[{\includegraphics[width=1in,height=1.25in,clip,keepaspectratio]{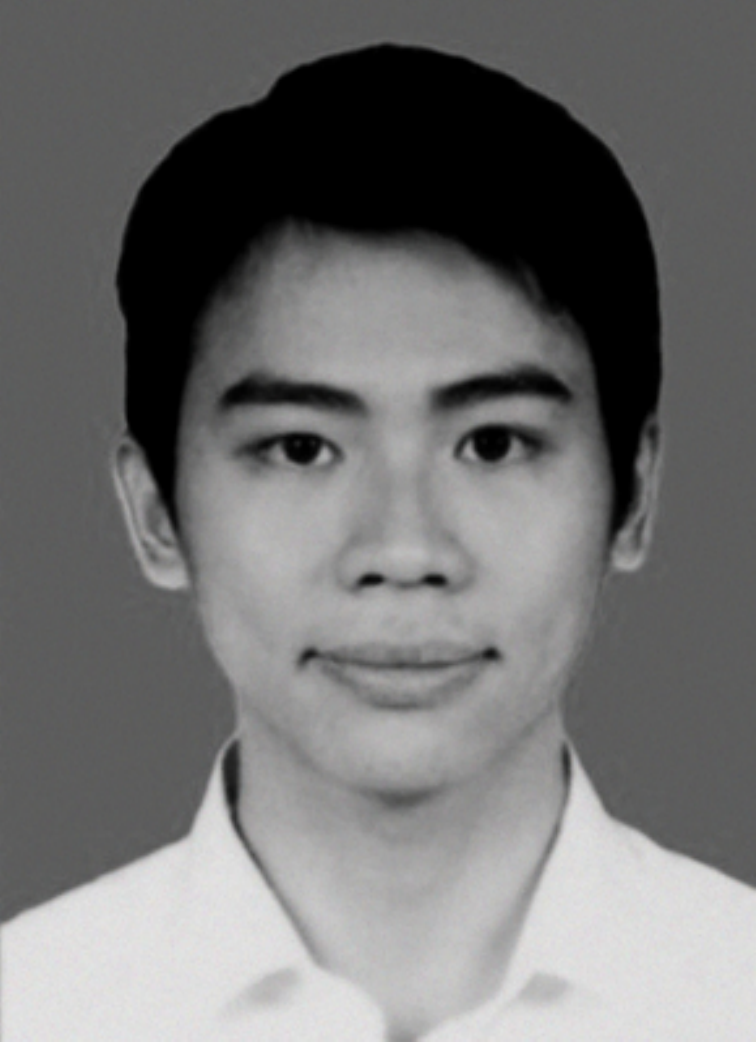}}]{Hefeng Wu}
received B.S. and Ph.D. degrees in computer science and technology from Sun Yat-sen University, China, in 2008 and 2013, respectively. He is currently a research fellow with the School of Data and Computer Science, Sun Yat-sen University, Guangzhou, China, and is also with the School of Information Science and Technology, Guangdong University of Foreign Studies, China. His research interests include image/video analysis, computer vision, and machine learning.
\end{IEEEbiography}
\vspace{-20pt}
\begin{IEEEbiography}[{\includegraphics[width=1in,height=1.25in,clip,keepaspectratio]{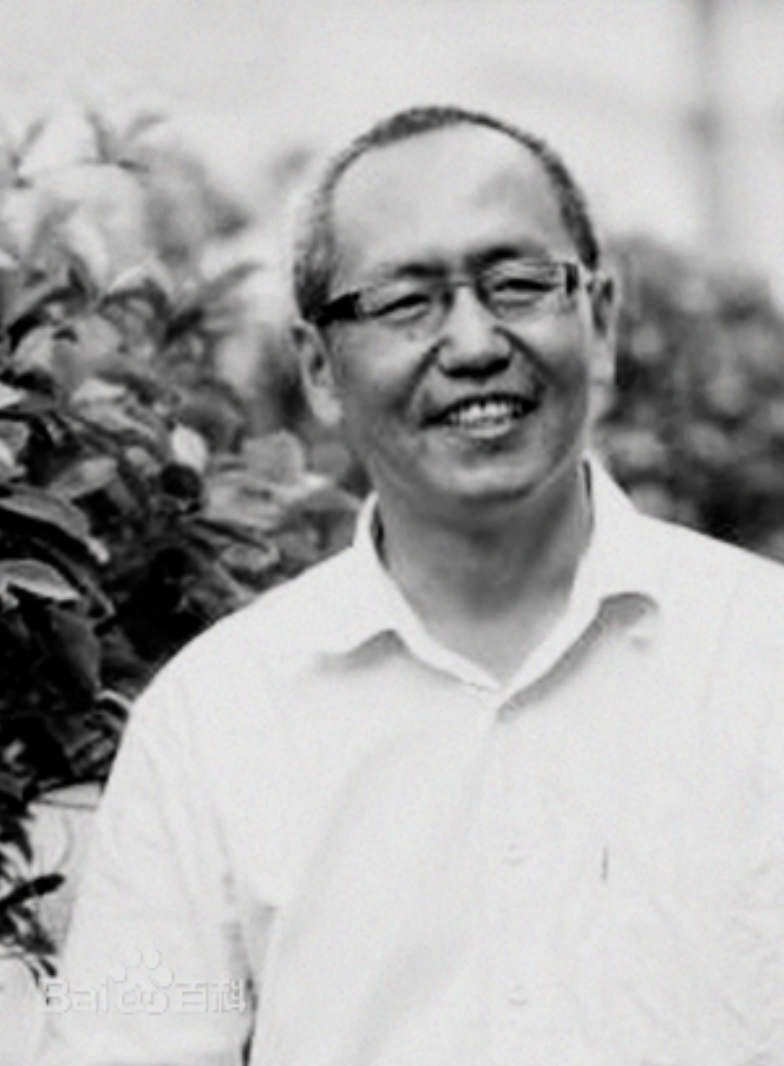}}]{Zhi Yang}
received his B.E. degree and Master's Degree from Lanzhou Technology University. He is a Professor at Sun Yat-sen University and the Executive Director of Guangdong Automation Society. His current research interests include information processing, automatic control technology and computer control technology.
\end{IEEEbiography}
\vspace{-20pt}
\begin{IEEEbiography}[{\includegraphics[width=1in,height=1.25in,clip,keepaspectratio]{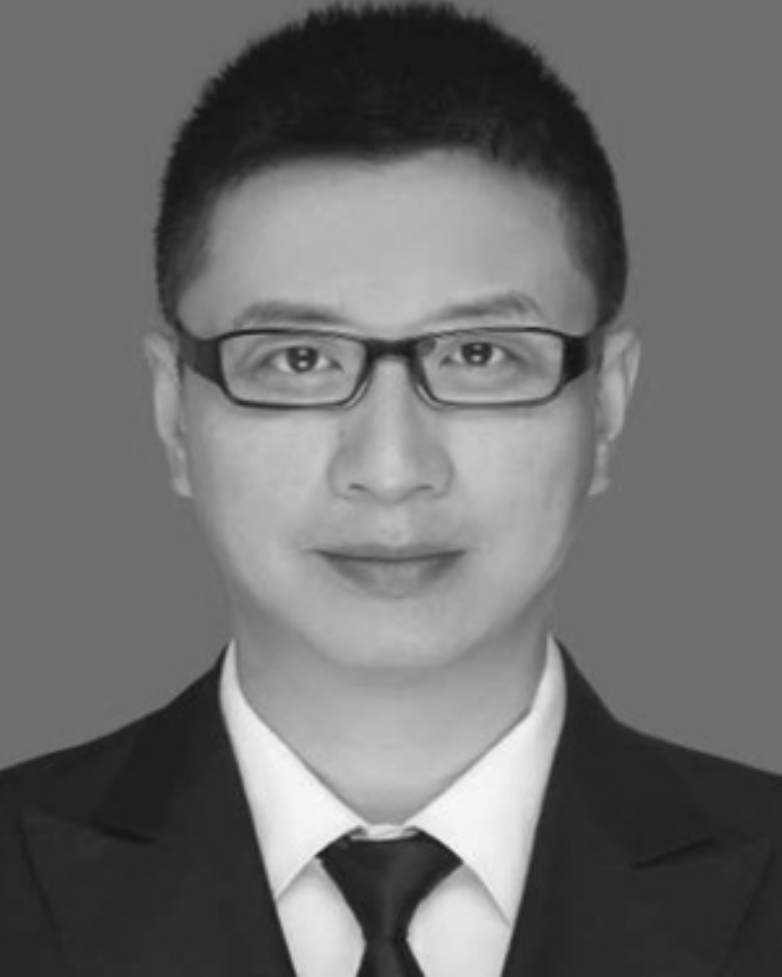}}]{Guangchun Luo }
(M'06) received the Ph.D. degree in Computer Science from the University of Electronic Science and Technology of China (UESTC), Chengdu, China, in 2004. He is currently a Full Professor with UESTC and serves as the Associate Dean of Computer Science and the Director of the Credible Cloud Computing and Big Data Laboratory at UESTC. He is also the ChangJiang Scholar Distinguished Professor appointed by the Ministry of Education of China. He has published over 70 journal and conference papers in his fields. His current research interests include computer networks, multimedia, cloud computing, and big data mining.
\end{IEEEbiography}

\begin{IEEEbiography}[{\includegraphics[width=1in,clip]{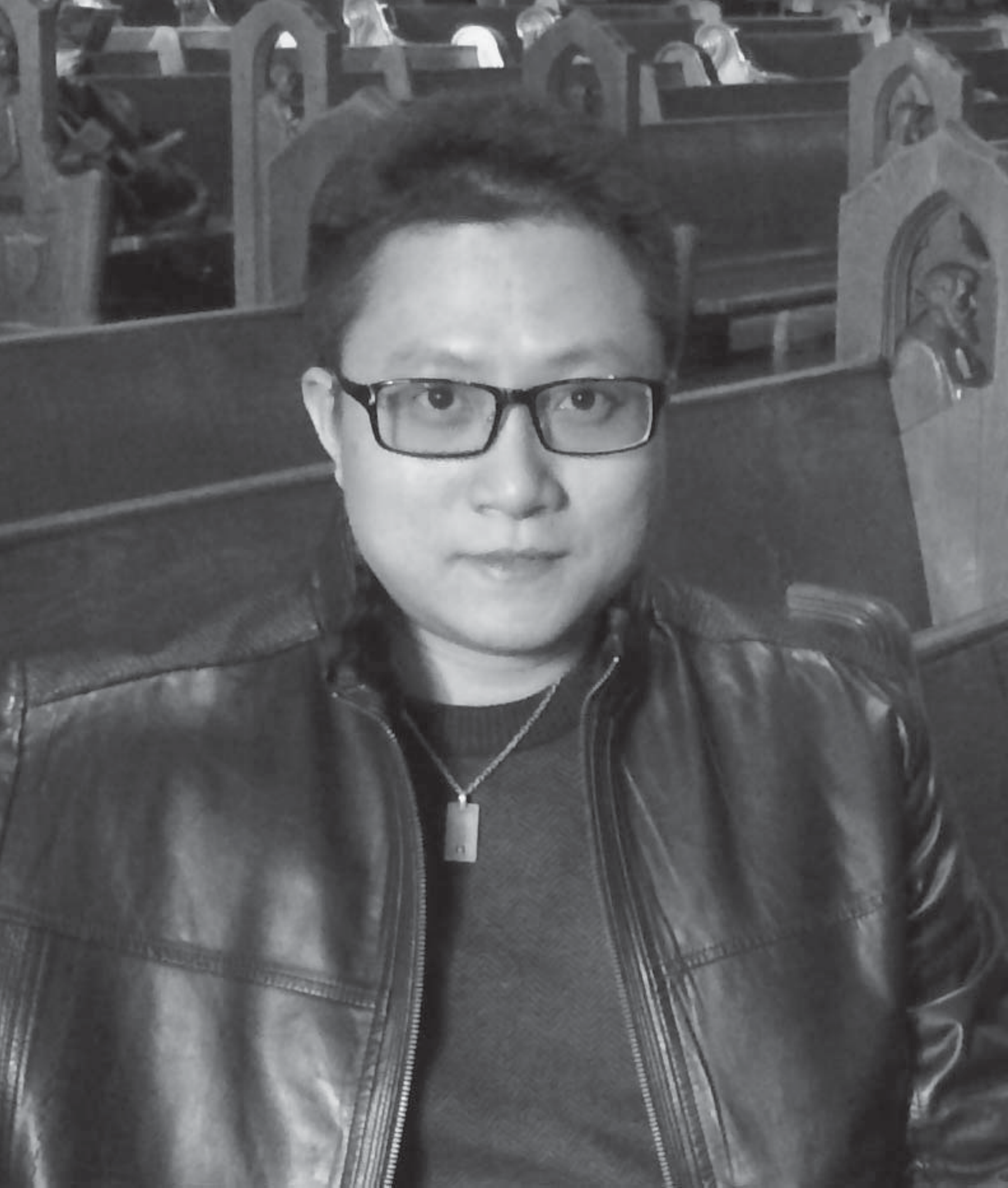}}]{Liang Lin} (M'09, SM'15) is the Executive Research Director of SenseTime Group Limited and a full Professor at Sun Yat-sen University. He is the Excellent Young Scientist of the National Natural Science Foundation of China. From 2008 to 2010, he was a Post-Doctoral Fellow at the University of California, Los Angeles. From 2014 to 2015, as a senior visiting scholar, he was with The Hong Kong Polytechnic University and The Chinese University of Hong Kong. He currently leads the SenseTime R\&D teams in developing cutting-edge and deliverable solutions in computer vision, data analysis and mining, and intelligent robotic systems. He has authored and coauthored more than 100 papers in top-tier academic journals and conferences (e.g., 10 papers in TPAMI/IJCV and 40+ papers in CVPR/ICCV/NIPS/IJCAI). He has been serving as an associate editor of IEEE Trans. on Human-Machine Systems, The Visual Computer and Neurocomputing. He served as the Area/Session Chair for numerous conferences such as ICME, ACCV, and ICMR. He was the recipient of the Best Paper Runner-Up Award at ACM NPAR 2010, the Google Faculty Award in 2012, the Best Paper Diamond Award at IEEE ICME 2017, and Hong Kong Scholars Award in 2014. He is a Fellow of IET.
\end{IEEEbiography}

\end{document}